\documentclass[journal]{IEEEtran}
\usepackage{cite}
\usepackage{graphicx}
\usepackage{url}
\usepackage{amsmath}
\usepackage[caption = false, font=footnotesize]{subfig}
\usepackage{float}
\usepackage{booktabs} % Allows the use of \toprule, \midrule and \bottomrule in tables for horizontal lines
\usepackage{amsfonts}

\usepackage{array}
\usepackage{color}
\usepackage{tabularx}
\usepackage{longtable}
\usepackage{tabu}
\usepackage[colorlinks,linkcolor=black]{hyperref}
\usepackage{multirow}
\newcolumntype{L}[1]{>{\raggedright\let\newline\\\arraybackslash\hspace{0pt}}m{#1}}
\newcolumntype{C}[1]{>{\centering\let\newline\\\arraybackslash\hspace{0pt}}m{#1}}
\newcolumntype{R}[1]{>{\raggedleft\let\newline\\\arraybackslash\hspace{0pt}}m{#1}}
\newcommand{\etal}{\textit{et al}.}
\newcommand{\ie}{\textit{i}.\textit{e}.}
\newcommand{\eg}{\textit{e}.\textit{g}.}

\usepackage{blkarray, bigstrut}

\definecolor{grey}{RGB}{130,130,130}
\definecolor{black}{RGB}{0,0,0}

\ifCLASSINFOpdf
\else
\fi

\begin{document}
\title{Uncertainty-Aware Blind Image Quality Assessment in the Laboratory and Wild}

%\author{Anonymous}
\author{Weixia~Zhang,~\IEEEmembership{Member,~IEEE,}
        Kede~Ma,~\IEEEmembership{Member,~IEEE,}
        Guangtao~Zhai,~\IEEEmembership{Senior Member,~IEEE,}
        and~Xiaokang~Yang,~\IEEEmembership{Fellow,~IEEE}% <-this % stops a space
\thanks{This work was supported in part by the National Natural Science Foundation of China under Grants 62071407 and 61901262, the CityU Start-up Grant (No. 7200630), and the CityU Strategic Research Grant (No. 7005560). }
\thanks{Weixia Zhang, Guangtao Zhai, and Xiaokang Yang are with the MoE Key Lab of Artificial Intelligence, AI Institute, Shanghai Jiao Tong University, Shanghai, China (e-mail: zwx8981@sjtu.edu.cn; zhaiguangtao@sjtu.edu.cn; xkyang@sjtu.edu.cn).}
\thanks{Kede Ma is with the Department of Computer Science, City University of Hong Kong, Kowloon, Hong Kong (e-mail: kede.ma@cityu.edu.hk).}
%\thanks{}
}


\markboth{}%
{Shell \MakeLowercase{\textit{et al.}}: Bare Demo of IEEEtran.cls for Journals}




\maketitle

\begin{abstract}
Performance of blind image quality assessment (BIQA) models has been significantly boosted by end-to-end optimization of feature engineering and quality regression. Nevertheless, due to the distributional shift between images simulated in the laboratory and captured in the wild, models trained on databases with synthetic distortions remain particularly weak at handling realistic distortions (and vice versa). To confront the cross-distortion-scenario challenge, we develop a \textit{unified} BIQA model and an approach of training it for both synthetic and realistic distortions.  We first sample pairs of images from individual IQA databases, and compute a probability that the first image of each pair is of higher quality. We then employ the fidelity loss to optimize a deep neural network for BIQA
over a large number of such image pairs. We also explicitly enforce a hinge constraint to regularize uncertainty estimation during optimization. Extensive experiments on six IQA databases show the promise of the learned method in blindly assessing image quality in the laboratory and wild. In addition, we demonstrate the universality of the proposed training strategy by using it to improve existing BIQA models.
\end{abstract}


% \fi
\IEEEpeerreviewmaketitle
\section{Introduction}\label{sec:intro}
\IEEEPARstart{W}{ith} the inelastic demand for processing massive Internet images, it is of paramount importance to develop computational image quality models to monitor, maintain, and enhance the perceived quality of the output images of various image processing systems~\cite{wang2006modern}. High degrees of consistency between model predictions and human opinions of image quality have been achieved in the full-reference regime, where distorted images are compared to their reference images of pristine quality~\cite{wang2003multiscale}. When such information is not available,
% or may not exist, such as for images captured in the wild~\cite{ciancio2011no},
no-reference or blind image quality assessment (BIQA) that relies solely on distorted images becomes more practical yet more challenging. Recently, deep learning-based BIQA models have experienced an impressive series of successes due to joint optimization of feature representation and quality prediction. However, these models remain particularly weak
at cross-distortion-scenario generalization~\cite{zhang2020blind}. That is, models trained on images simulated in the laboratory cannot deal with images captured in the wild. Similarly, models optimized for realistic distortions (\eg, sensor noise and poor exposure) do not work well for synthetic distortions (\eg, Gaussian blur and JPEG compression).

% \begin{figure}
%   \centering
%   \includegraphics[width=.5\textwidth]{figs/challenges}
%   \caption{Two challenges in developing a unified BIQA model to handle diverse distortion scenarios within databases with different perceptual scales.}\label{fig:challenge}
% \end{figure}

\begin{figure*}[t]
    \centering
    \captionsetup{justification=centering}
    \subfloat[]{\includegraphics[width=0.33\textwidth]{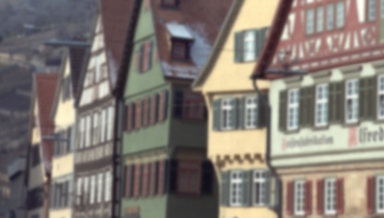}}\hskip.2em
    \subfloat[]{\includegraphics[width=0.33\textwidth]{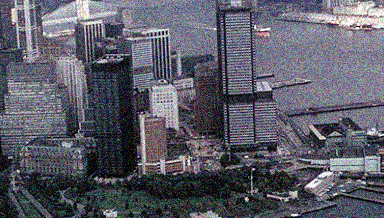}}\hskip.2em
    \subfloat[]{\includegraphics[width=0.33\textwidth]{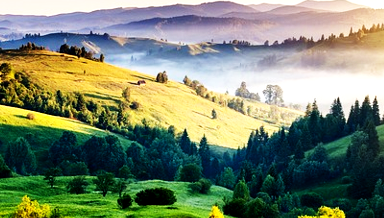}}
    \vspace{-.5em}
    \subfloat[]{\includegraphics[width=0.33\textwidth]{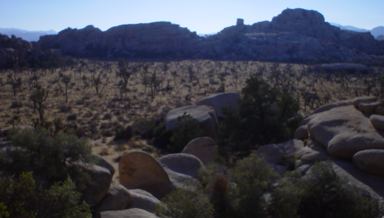}}\hskip.2em
    \subfloat[]{\includegraphics[width=0.33\textwidth]{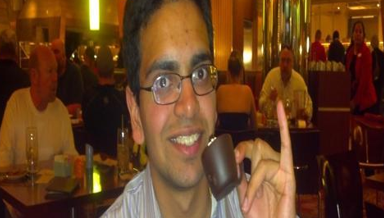}}\hskip.2em
    \subfloat[]{\includegraphics[width=0.33\textwidth]{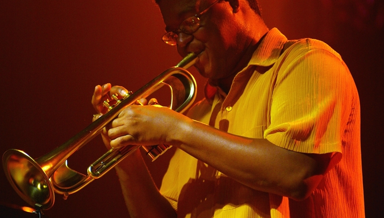}}
  \caption{Images with approximately the same linearly re-scaled MOS exhibit drastically different perceptual quality. If the human scores are in the form of DMOSs, we first negate the values followed by linear re-scaling. These are sampled from (a) LIVE~\cite{sheikh2006statistical}, (b) CSIQ~\cite{larson2010most}, (c) KADID-10K~\cite{lin2019kadid}, (d) BID~\cite{ciancio2011no}, (e) LIVE Challenge~\cite{ghadiyaram2016massive}, and (f) KonIQ-10K~\cite{hosu2020koniq}. It is not hard to observe that image (f) has clearly superior quality than the other five. All images are cropped for better visibility.}
\label{fig:quality_comparison}
\end{figure*}

A seemingly straightforward method of adapting to the distributional shift between synthetic and realistic distortions is to directly combine multiple IQA databases for training. However, existing databases have different perceptual scales due to differences in subjective testing methodologies. For example, the CSIQ database~\cite{larson2010most} used a multiple stimuli absolute category rating in a well-controlled laboratory environment, with difference
mean opinion scores (DMOSs) in the range of $[0,1]$, whereas the LIVE Challenge Database used a single stimulus continuous quality rating in an unconstrained crowdsourcing platform, with MOSs in the range of $[0,100]$. This means that a separate subjective experiment on images sampled from each database is required for perceptual scale realignment~\cite{sheikh2006statistical, larson2010most}. To make this point more explicit, we linearly re-scaled the subjective scores of each of six databases~\cite{sheikh2006statistical,larson2010most,lin2019kadid,ciancio2011no,ghadiyaram2016massive,hosu2020koniq} to $[0,100]$, with a larger value indicating higher quality. Fig.~\ref{fig:quality_comparison} shows the sample images that have approximately the same re-scaled MOS.
As expected, they appear to have drastically different perceptual quality. A more promising design methodology for \textit{unified} BIQA is to build a prior probability model for natural undistorted images as the reference distribution, to which a test distribution of the distorted image can be compared. The award-winning BIQA model - NIQE~\cite{mittal2013making} is a specific instantiation of this idea, but is only capable of handling a small number of synthetic distortions.

In addition to training BIQA models with (D)MOSs, there is another type of supervisory signal - the variance of human opinions, which we believe is beneficial for BIQA, but has not been fully explored. Generally, humans tend to give more consistent ratings (\ie, smaller variances) to images at the two ends of the quality range, while assessing images in the mid-quality range with less certainty (see Fig.~\ref{fig:gt_curve}). Therefore, it is reasonable to assume image quality models to behave similarly. Moreover, previous BIQA methods~\cite{wu2018blind}
have enjoyed the benefits of modeling the uncertainty of quality prediction for subsequent applications.

In this paper, we take steps toward developing unified uncertainty-aware BIQA models for both synthetic and realistic distortions. Our contributions include:
\begin{itemize}
    \item A training strategy that allows differentiable BIQA models to be trained on multiple IQA database (of different distortion scenarios) simultaneously. In particular, we first sample and combine pairs of images within each database. For each pair, we leverage the human-rated (D)MOSs and variances to compute a probability value that one image is of better perceptual quality as the supervisory signal. The resulting training set bypasses additional subjective testing for perceptual scale realignment. We then use a pairwise learning-to-rank algorithm with the fidelity loss~\cite{tsai2007frank} to drive the learning of computational models for BIQA.
    \item A regularizer that enforces a hinge constraint on the learned uncertainty using the variance of human opinions as guidance. This enables BIQA models to mimic the uncertain aspects of humans when performing the quality assessment task.
    \item A Unified No-reference Image Quality and Uncertainty Evaluator (UNIQUE) based on a deep neural network (DNN) that \textcolor{black}{performs favorably against} state-of-the-art BIQA models on six IQA databases (see Table~\ref{tab:database}) covering both synthetic and realistic distortions. We also verify its generalizability in a challenging cross-database setting and via the group maximum differentiation (gMAD) competition~\cite{ma2020group}.
    % \item A Unified No-reference Image Quality and Uncertainty Evaluator (UNIQUE) based on a deep neural network (DNN) that significantly outperforms state-of-the-art BIQA models on six IQA databases (see Table~\ref{tab:database}) covering both synthetic and realistic distortions. We also verify its generalizability in a challenging cross-data setting and via the group maximum differentiation (gMAD) competition methodology~\cite{ma2020group}.
\end{itemize}

\begin{table*}[t]
  \centering
  \caption{Comparison of human-rated IQA databases. MOS stands for mean opinion score. DMOS is inversely proportional to MOS }\label{tab:database}
  \begin{tabular}{l|ccccc}
      \toprule
        {Database} & \# of Images & Scenario & Annotation & Range & Subjective Testing Methodology\\
     \hline
        LIVE~\cite{sheikh2006statistical} & 779 & Synthetic & DMOS, Variance & $[0, 100]$ & Single stimulus continuous quality rating\\
        CSIQ~\cite{larson2010most} & 866 & Synthetic & DMOS, Variance & $[0, 1]$ & Multi stimulus absolute category rating\\
        %TID2013~\cite{ponomarenko2013color} & synthetic & MOS, std & [0, 9] & Two-alternative forced choice\\
        KADID-10K~\cite{lin2019kadid} & 10,125 & Synthetic & MOS, Variance & $[1, 5]$ & Double stimulus absolute category rating with crowdsourcing\\
    \hline
        BID~\cite{ciancio2011no} & 586 & Realistic & MOS, Variance & $[0, 5]$ & Single stimulus continuous quality rating\\
        LIVE Challenge~\cite{ghadiyaram2016massive} & 1,162 & Realistic & MOS, Variance & $[0, 100]$ & Single stimulus continuous quality rating with crowdsourcing\\
        KonIQ-10K~\cite{hosu2020koniq} & 10,073 & Realistic & MOS, Variance & $[1, 5]$ & Single stimulus absolute category rating with crowdsourcing\\
     \bottomrule
   \end{tabular}
\end{table*}

\textbf{\begin{table*}[t]
  \small
  \centering
  \caption{Summary of ranking-based BIQA models. DS: distortion specification characterized by distortion parameters. FR: full-reference IQA model predictions}\label{tab:rank_comparison}
  \begin{tabular}{l|ccccccc}
      \toprule
        {Model} & RankIQA~\cite{liu2017rankiqa} & DB-CNN~\cite{zhang2020blind} & dipIQ~\cite{ma2017dipiq} & Ma19~\cite{ma2019blind} & Gao15~\cite{gao2015learning} & UNIQUE\\
     \hline
        Source & DS & DS & FR & FR & (D)MOS & (D)MOS+Variance\\
        Scenario & Synthetic & Synthetic & Synthetic & Synthetic & Synthetic & Synthetic+Realistic\\
        Annotation & Binary & Categorical & Binary & Binary & Binary & Continuous\\
        Loss Function & Hinge variant & Cross entropy & Cross entropy & Cross entropy variant & Hinge & Fidelity+Hinge\\
        Ranking Stage & Pre-training & Pre-training & Prediction & Prediction & Prediction & Prediction\\
       % Uncertainty Supervision & $\times$ & $\times$  & $\times$  & $\times$  & $\times$  & $\surd$\\
%         & DS & Syn & RH\\
%         & DS & Syn & CE\\
%         & FR & Syn & Custom\\
%         & FR & Syn & CE\\
%         & FR & Syn & CE\\
%         & M & Syn & MKLGL\\
%         & MS & Syn + Real & Fidelity\\
     \bottomrule
   \end{tabular}
\end{table*}}

\section{Related Work}
In this section, we give a review of existing BIQA models over the last two decades.
\subsection{BIQA as Regression}
Early attempts at BIQA were tailored to specific synthetic distortions~\cite{zhai2020perceptual}, such as  JPEG compression~\cite{wang2002no} and JPEG2000 compression~\cite{marziliano2004perceptual}. Later models aimed for general-purpose BIQA~\cite{moorthy2011blind,ye2012unsupervised,mittal2012no,xu2016blind,ghadiyaram2017perceptual}, with the underlying assumption that statistics extracted from natural images are highly regular~\cite{simoncelli2001natural}
and distortions will break such statistical regularities. Based on natural scene statistics (NSS), a quality prediction function can be learned using standard supervised learning tools. Of particular interest is NIQE~\cite{mittal2013making}, which is arguably the first unified BIQA model with the goal of capturing arbitrary distortions. However, the NSS model used in NIQE is not sensitive to image ``unnaturalness'' introduced by realistic distortions~\cite{mittal2013making}. Zhang~\textit{et al.}~\cite{zhang2015feature} extended NIQE~\cite{mittal2013making} by exploiting a  more powerful set of  NSS for local quality prediction. However, the generalization to realistic distortions is still limited.

Joint optimization of feature engineering and quality regression enabled by deep learning has significantly advanced the field of BIQA in recent years. The apparent conflict between the small number of subjective ratings and the large number of learnable model parameters may be alleviated in three ways. The first method is transfer learning~\cite{zeng2018blind}, which directly fine-tunes pre-trained DNNs for object recognition.
\textcolor{black}{Recently, Zhu~\etal~\cite{zhu2020metaiqa} explored meta-learning as an advanced form of transfer learning in BIQA.} The second method is patch-based training, which assigns a local quality score to an image patch transferred from the corresponding global quality score~\cite{kang2014convolutional, bosse2016deep}. The third method is quality-aware pre-training, which automatically generates a large amount of labeled data by exploiting specifications of distortion processes or quality estimates of full-reference models~\cite{Ma2018End,liu2017rankiqa,zhang2020blind}.  Despite impressive correlation numbers on individual databases of either synthetic or realistic distortions, DNN-based BIQA models are vulnerable to cross-distortion-scenario generalization, and can also be easily falsified in the gMAD competition~\cite{wang2020active}.

\subsection{BIQA as Ranking}
There are also methods that cast BIQA as a learning-to-rank problem~\cite{hu2019pairwise}, where \textit{relative} ranking information can be obtained from distortion specifications~\cite{liu2017rankiqa, Ma2018End, zhang2020blind}, full-reference IQA models~\cite{ma2017dipiq,ma2019blind}, and human data~\cite{gao2015learning}. Liu~\textit{et al.}~\cite{liu2017rankiqa} and Zhang~\textit{et al.}~\cite{zhang2020blind} inferred discrete ranking information from images of the same content and distortion  but at different levels for BIQA model pre-training. Different from~\cite{liu2017rankiqa,zhang2020blind}, the proposed UNIQUE
explores continuous ranking information from (D)MOSs and variances in the stage of final quality prediction. Ma~\textit{et al.}~\cite{ma2017dipiq,ma2019blind} extracted binary ranking information from full-reference IQA methods to guide the optimization of BIQA models.
Since full-reference methods can only be applied to synthetic distortions, where the reference images are available, it is not trivial to extend the methods in~\cite{ma2017dipiq,ma2019blind} to realistic distortions. The closest work to ours is due to Gao~\textit{et al.}~\cite{gao2015learning}, who computed binary rankings from MOSs. However, they neither performed end-to-end optimization of BIQA nor explored the idea of combining multiple IQA databases via pairwise rankings. As a result, their method only achieves reasonable performance on a limited number of synthetic distortions. UNIQUE takes a step further to be uncertainty-aware, learning from human behavior when evaluating image quality. We summarize ranking-based BIQA methods in Table~\ref{tab:rank_comparison}.

\begin{figure*}
  \centering
  \includegraphics[width=0.95\textwidth]{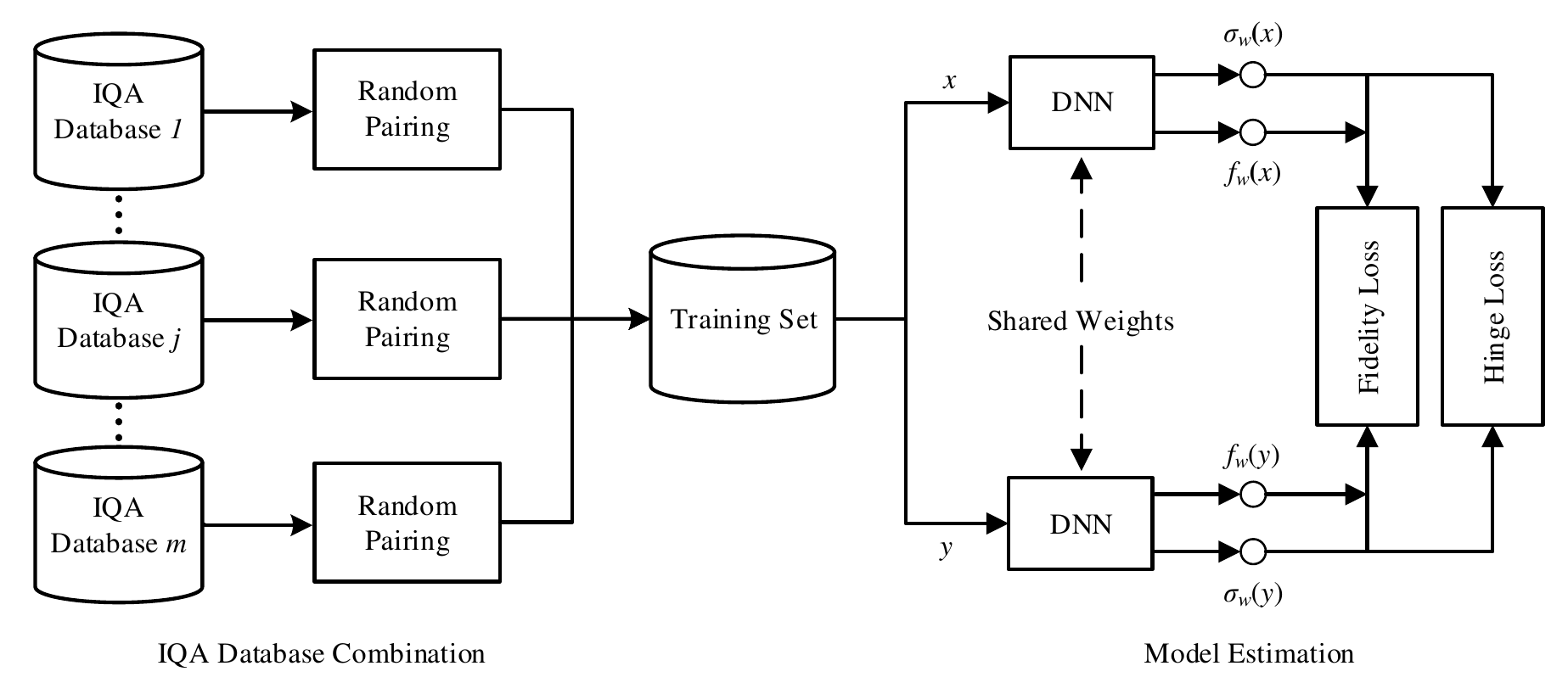}
  \caption{Illustration of the proposed training strategy, which involves two steps: IQA database combination and pairwise learning-to-rank model estimation. The training pairs are randomly sampled within each IQA database and then combined. The optimization is driven by the fidelity and hinge losses.}\label{fig:framework}
\end{figure*}

\subsection{Uncertainty-Aware BIQA}
Learning uncertainty is helpful to understand and analyze model predictions. In Bayesian machine learning, uncertainty may come from two parts: one inherent in the data (\ie, data/aleatoric uncertainty) and the other in the learned parameters (\ie, model/epistemic uncertainty)~\cite{kendall2017what}. In the context of BIQA, Huang~\textit{et al.}~\cite{huang2019convolutional}
modeled the uncertainty of patch quality to alleviate the label noise problem in patch-based training. Wu~\textit{et al.}~\cite{wu2018blind} employed a sparse Gaussian process for quality regression, where the data uncertainty can be jointly estimated without supervision. In contrast, UNIQUE assumes the Thurstone's model~\cite{thurstone1927law}, and learns the data uncertainty with direct supervision, aiming for an effective BIQA model with a probabilistic interpretation.
%Kendall~\textit{et al.}~\cite{kendall2017what} exploit both aleatoric uncertainty and epistemic uncertainty for semantic segmentation and pixel-wise depth regression.

\section{Training UNIQUE}
\label{sec:uiqe}
In this section, we first present the proposed training strategy, consisting of IQA database combination and pairwise learning-to-rank model estimation (see Fig.~\ref{fig:framework}). We then describe the details of the UNIQUE model for unified uncertainty-aware BIQA.

\subsection{IQA Database Combination}\label{training_set}
Our goal is to combine $m$ IQA databases for training while avoiding additional subjective experiments for perceptual scale realignment. To achieve this, we first  randomly sample $n_j$ pairs of images $\lbrace( x^{j}_i,y^{j}_i)\rbrace_{i=1}^{n_j}$ from the $j$-th database. For each image pair $(x, y)$, we infer relative ranking information from the corresponding MOSs and variances. Specifically, under the Thurstone's model~\cite{thurstone1927law}, we assume that the true perceptual quality $q(x)$ of image $x$ follows a Gaussian distribution with mean $\mu(x)$ and variance $\sigma^2(x)$ collected via subjective testing. Assuming the variability of quality across images is uncorrelated, the quality difference, $q(x)-q(y)$, is also Gaussian with mean $\mu(x) - \mu(y)$ and variance $\sigma^{2}(x) + \sigma^{2}(y)$. The probability that $x$ has higher perceptual quality than $y$ can be calculated from the Gaussian cumulative distribution function $\Phi(\cdot)$:
\begin{align}\label{eq:difference}
p(x,y)= \Pr({q(x) \ge q(y)}) = \Phi\left(\frac{\mu(x) - \mu(y)}{\sqrt{\sigma^{2}(x) + \sigma^{2}(y)}}\right).
\end{align}
Combining pairs of images from $m$ databases, we are able to build a training set $\mathcal{D}= \lbrace\lbrace(x^j_i,  y^j_i),p^j_i\rbrace_{i=1}^{n_j}\rbrace_{j=1}^{m}$. Our database combination approach allows future IQA databases to be added with essentially no cost.

\subsection{Model Estimation}\label{L2R}
Given $\mathcal{D}$ as the training set, we aim to learn two differentiable functions $f_{w}(\cdot)$ and $\sigma_{w}(\cdot)$, parameterized by a vector $w$, which accept an image of arbitrary input size, and compute the quality score and the uncertainty, respectively. Similar in Section~\ref{training_set}, we assume the true perceptual quality $q(x)$ obeys a Gaussian distribution with mean and variance now estimated by $f_w(x)$ and $\sigma_w^2(x)$, respectively. The probability of preferring $x$ over $y$ perceptually is
\begin{align}\label{eq:difference2}
p_w(x,y) = \Phi\left(\frac{f_{w}(x) - f_{w}(y)}{\sqrt{\sigma_{w}^{2}(x) + \sigma_{w}^{2}(y)}}\right).
\end{align}
It remains to specify a similarity measure between the probability distributions $p(x,y)$ and $p_w(x,y)$ as the objective for model estimation. In machine learning, cross-entropy may be the de facto measure for this purpose, but has several drawbacks~\cite{tsai2007frank}. First, the minimum of the cross-entropy loss  is not exactly zero, except for the ground truth $p(x,y)=0$ and $p(x,y)=1$. This may hinder the learning of image pairs with $p(x,y)$ close to $0.5$. Second, the cross-entropy loss is unbounded from above, which may over-penalize some hard training examples, therefore biasing the learned models. To resolve these issues, we choose the fidelity loss~\cite{tsai2007frank}, originated from quantum physics~\cite{birrell1984quantum}, as the similarity measure:
\begin{align}\label{eq:fidelity}
\ell_F(x, y, p;w)
= 1& - \sqrt{p(x,y)p_w(x,y)}  \nonumber \\
&-\sqrt{(1-p(x,y))(1-p_w(x,y))}.
\end{align}
% The gradient of $\ell(x, y, r;w)$ with respect to the parameter vector $w$ is computed by
% \begin{align}\label{eq:loss_derivations}
% \frac{\partial{\ell(x, y, p;w)}}{\partial{w}}
% =& \frac{1}{2}\frac{\partial{p_{w}(x,y)}}{\partial{w}}\Bigg[-\frac{p(x,y)}{\sqrt{p(x,y)p_{w}(x,y)}}\nonumber \\
% &+ \frac{1-p(x,y)}{\sqrt{(1 - p(x,y)(1-p_{w}(x,y))}}\Bigg]
% \end{align}
Joint estimation of image quality and uncertainty will introduce scaling ambiguity. More precisely, if we make the scaling $f_w(\cdot)\rightarrow \alpha f_w(\cdot)$ and $\sigma_w(\cdot) \rightarrow \alpha\sigma_w(\cdot)$, the probability $p_w(x,y)$ given by Eq.~\eqref{eq:difference2} is unchanged. Our preliminary results~\cite{zhang2020learning} showed that the learned $\sigma_w(x)$ by optimizing Eq.~\eqref{eq:fidelity} solely  neither resembles
any aspects of human behavior in BIQA, nor reveals new statistical properties of natural images. To resolve the scaling ambiguity and provide with $\sigma_w(x)$  direct supervision, we enforce a regularizer of $\sigma_w(x)$ by taking advantage of the ground truth  $\sigma(x)$. Note that the ground truth \textcolor{black}{standard deviations} (stds) across IQA databases are not comparable, which prevents the use of their absolute values. Similarly, for each pair $(x,y)$, we infer
a binary label $t$ for uncertainty learning, where $t=1$ if $\sigma(x) \ge \sigma(y)$ and $t=-1$ otherwise. We  define the regularizer using  the hinge loss:
\begin{align}\label{eq:margin_ranking}
\ell_{H}(x,  y, t; w) = \max\left(0, \xi-t(\sigma_w(x)  - \sigma_w(y) )\right),
\end{align}
where the margin $\xi$ specifies the scale for BIQA models to work with. The augmented training set becomes $\mathcal{D}= \lbrace\lbrace(x^j_i,  y^j_i),p^j_i, t^j_i\rbrace_{i=1}^{n_j}\rbrace_{j=1}^{m}$.
During training, we sample a mini-batch $\mathcal{B}$ from $\mathcal{D}$ in each iteration, and use stochastic gradient descent to update the parameter vector $w$ by minimizing the following empirical loss:
\begin{align}\label{eq:loss1}
\ell(\mathcal{B}; w) =\frac{1}{\vert\mathcal{B}\vert} \sum_{\{(x, y), p, t\}\in \mathcal{B}}\ell_{F}( x,  y,p; w) + \lambda\ell_{H}(x, y,t; w),
\end{align}
% \begin{align}\label{eq:loss1}
% \ell(\mathcal{B}; w) =\frac{\sum_{\{(x, y), p, r\}\in \mathcal{B}}\ell_{F}( x,  y,p; w) + \lambda\ell_{H}(x, y,r; w)}{\vert\mathcal{B}\vert},
% \end{align}
where $\vert\mathcal{B}\vert$ denotes the cardinality of $\mathcal{B}$ and $\lambda$ trades off the two terms.
\begin{figure*}[t]
    \centering
    \captionsetup{justification=centering}
    \subfloat[LIVE]{\includegraphics[width=0.33\textwidth]{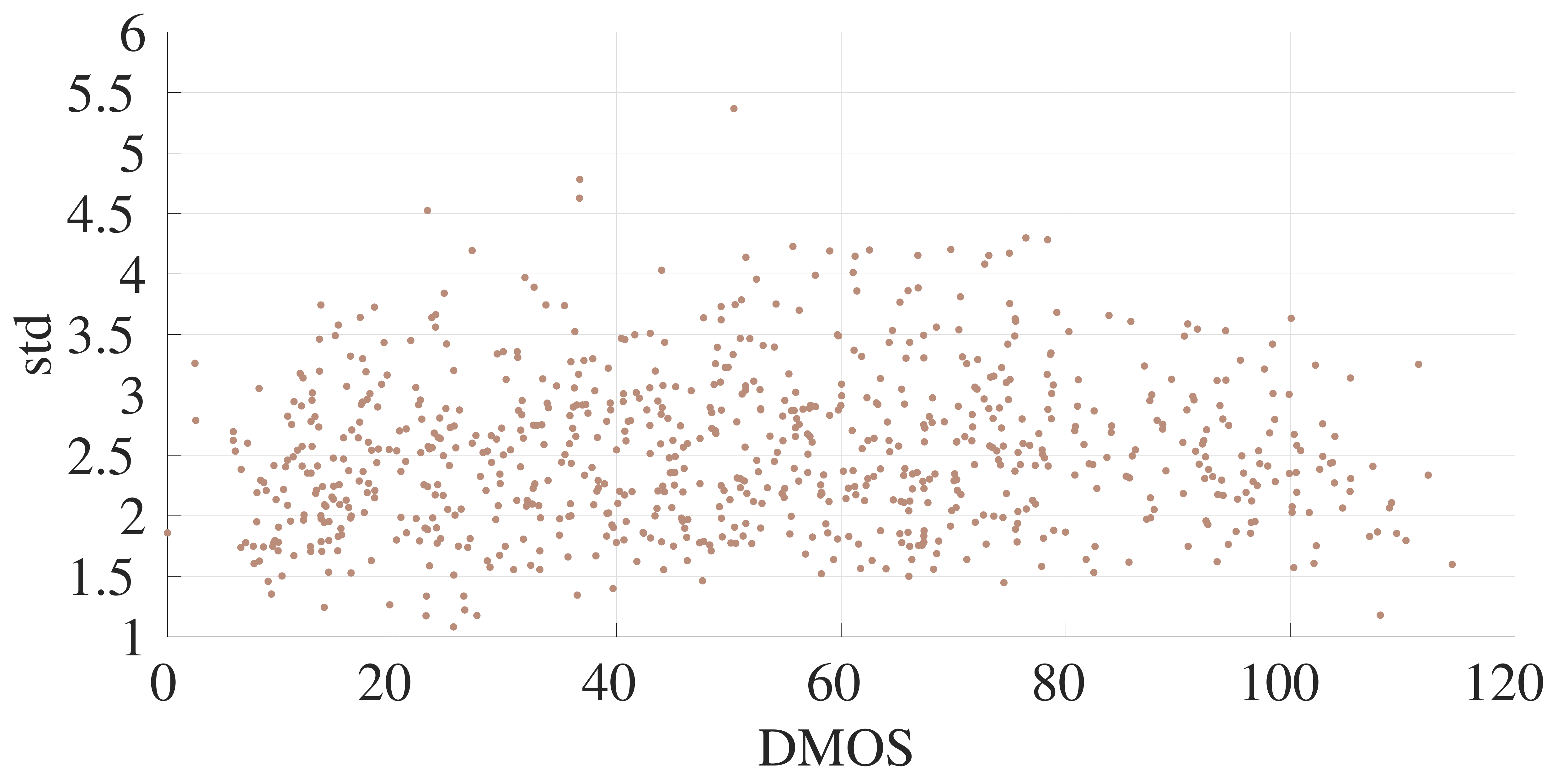}}
    \subfloat[CSIQ]{\includegraphics[width=0.33\textwidth]{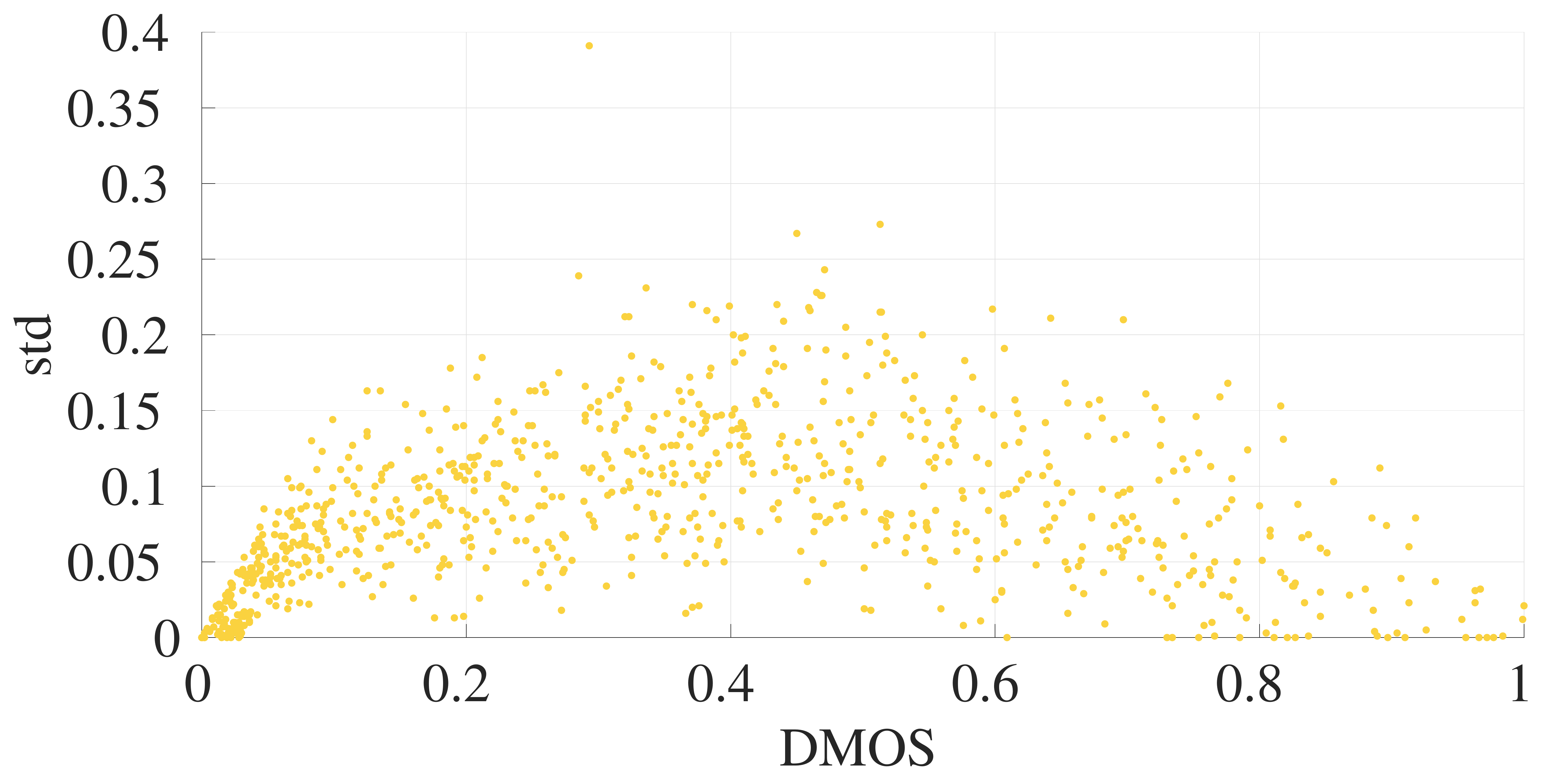}}
    \subfloat[KADID-10K]{\includegraphics[width=0.33\textwidth]{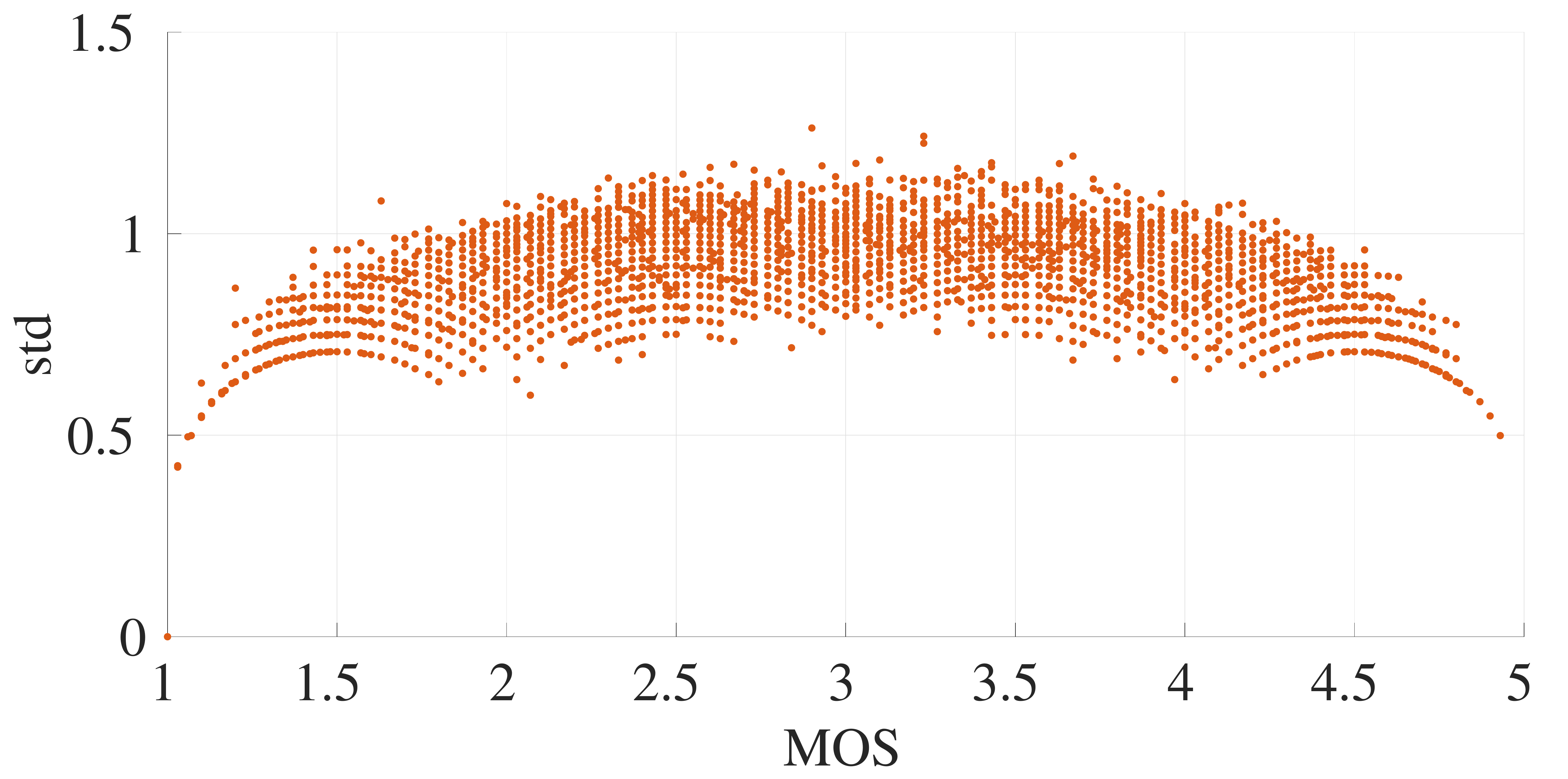}}
    \vspace{0em}
    \subfloat[BID]{\includegraphics[width=0.33\textwidth]{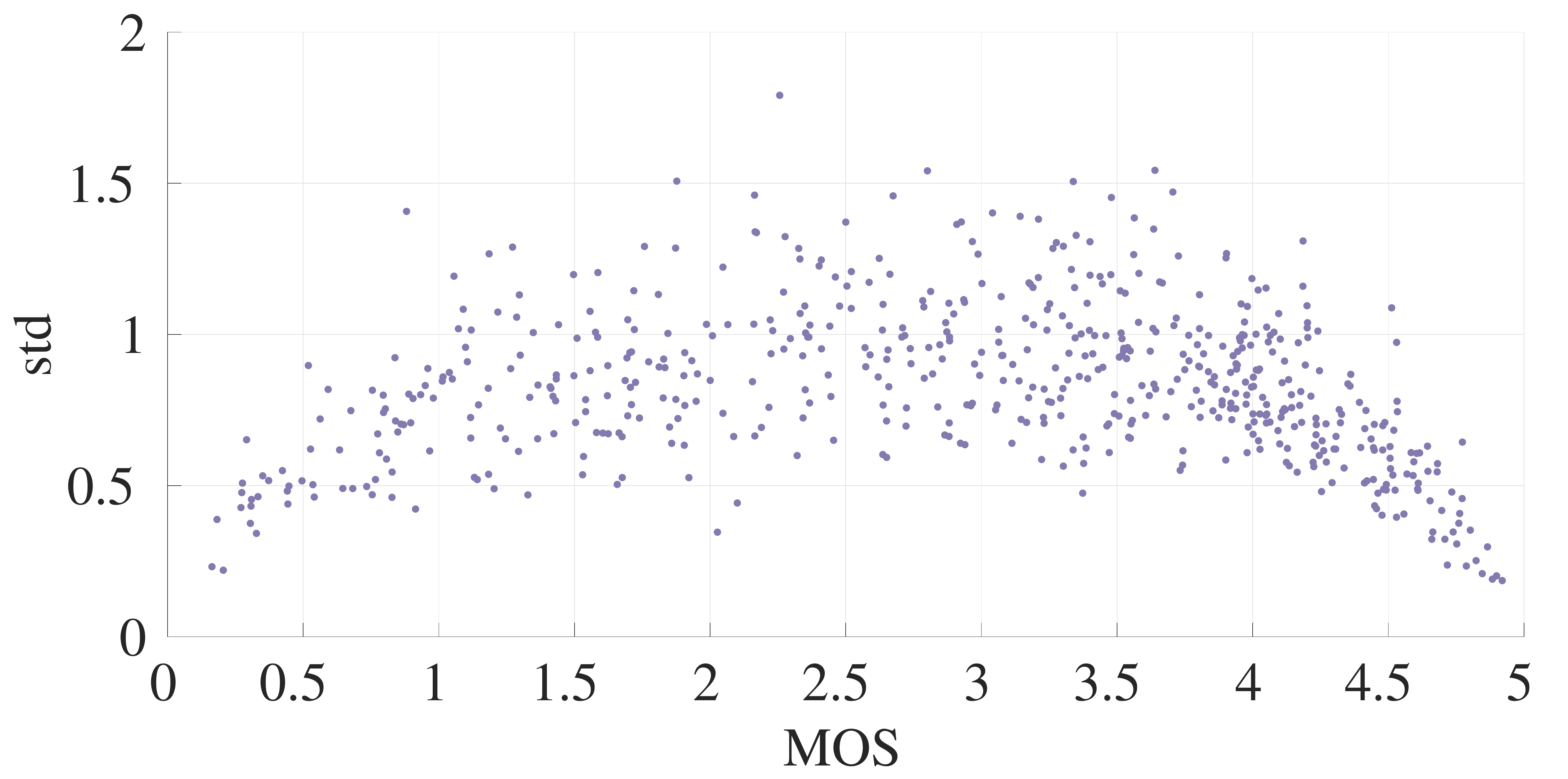}}
    \subfloat[LIVE Challenge]{\includegraphics[width=0.33\textwidth]{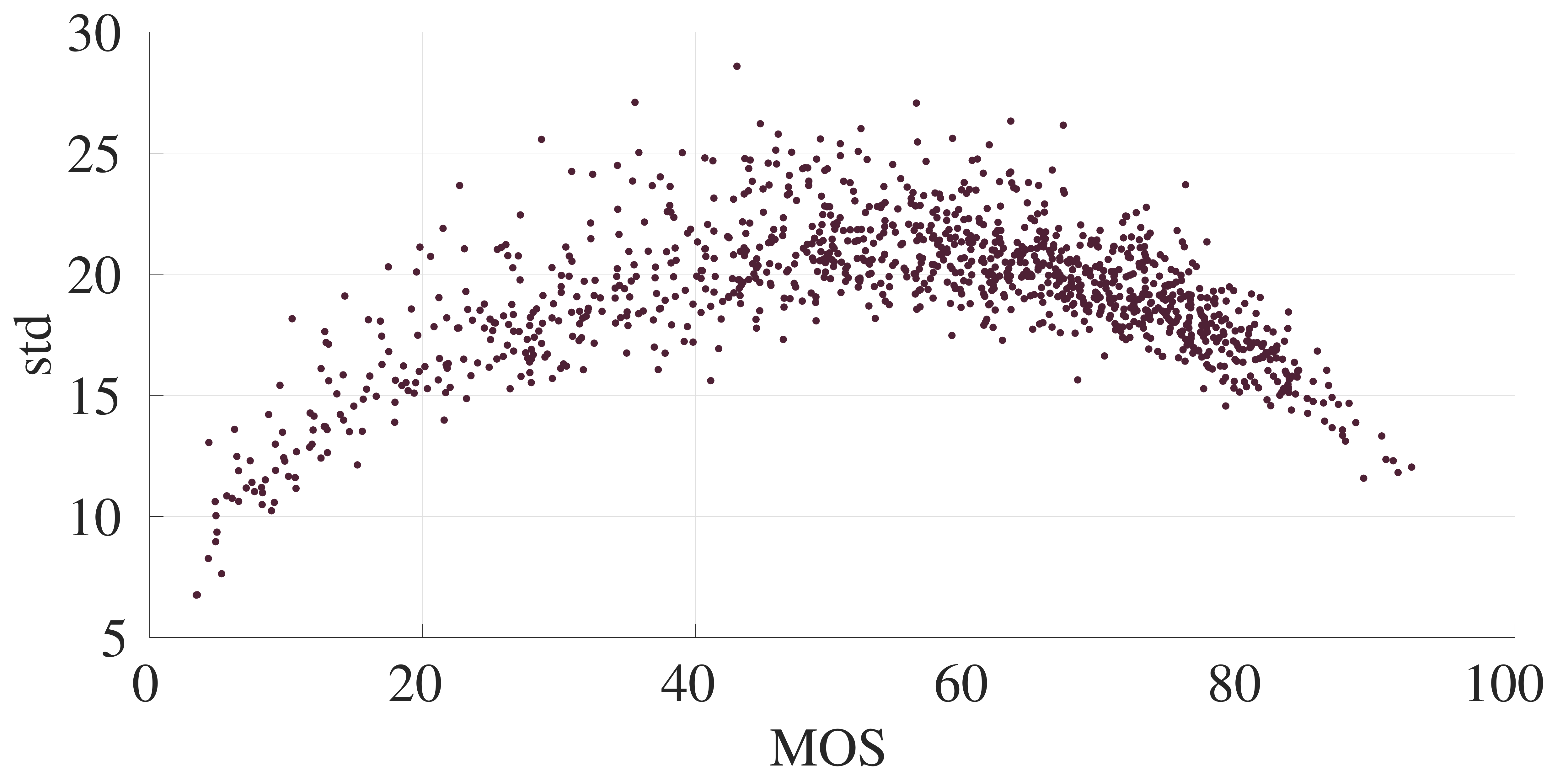}}
    \subfloat[KonIQ-10K]{\includegraphics[width=0.33\textwidth]{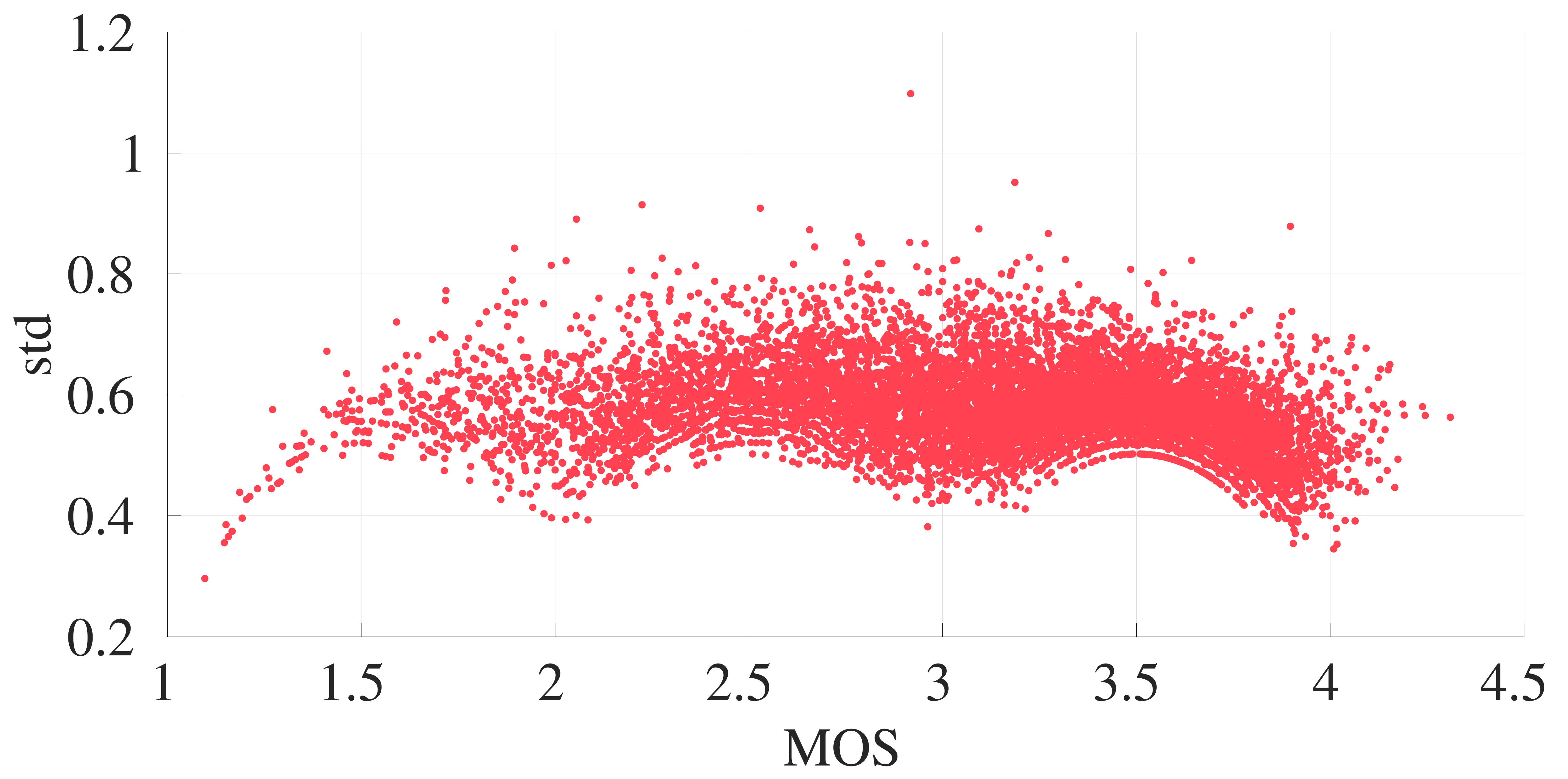}}
    %\vspace{-.2cm}
  \caption{Scatter plots of means against stds of human quality opinions of images from six IQA databases, including (a) LIVE~\cite{sheikh2006statistical}, (b) CSIQ~\cite{larson2010most}, (c) KADID-10K~\cite{lin2019kadid}, (d) BID~\cite{ciancio2011no}, (e) LIVE Challenge~\cite{ghadiyaram2016massive}, and (f) KonIQ-10K~\cite{hosu2020koniq}. There is a clear trend that humans are more consistent (\ie, confident) in making predictions of low-quality and high-quality images than mid-quality images, giving rise to arch-like shapes.}
\label{fig:gt_curve}
\end{figure*}

\subsection{Specification of UNIQUE}\label{subsec:ranknet}
We use ResNet-34~\cite{he2016deep} as the backbone of UNIQUE due to its good balance between model complexity and capability. Similar to Siamese neural networks~\cite{bromley1994signature}, the pairwise learning-to-rank framework composed of two streams is shown in Fig.~\ref{fig:framework}. Each stream is implemented by a DNN, consisting of a stage of convolution, batch normalization~\cite{ioffe2015batch}, ReLU nonlinearity, and max-pooling, followed by four residual blocks (see Table~\ref{table:backbone}). To generate a fixed-length image representation and summarize higher-order spatial statistics, we replace the first-order average pooling with a second-order bilinear pooling, which has been empirically proven useful in object recognition~\cite{lin2015bilinear} and BIQA~\cite{zhang2020blind}. We flatten the spatial dimensions of the feature representation after the last convolution to obtain $z \in \mathbb{R}^{s \times c}$, where $s$ and $c$ denote the spatial and channel dimensions, respectively. The bilinear pooling can be defined as
\begin{align}\label{eq:bp}
\bar{z} = z^T z.
\end{align}
We further flatten $\bar{z}\in \mathbb{R}^{c\times c}$, and append a fully connected layer with two outputs to represent $f_w(x)$ and $\sigma_w(x)$, respectively. The network parameters of the two streams are shared during the entire optimization process.

\newcommand{\blockc}[3]{\multirow{3}{*}{\(\left[\begin{array}{c}\text{3$\times$3, #1, stride #2}\\[-.1em] \text{3$\times$3, #1, stride #2} \end{array}\right]\)$\times$#3}
}

\newcommand{\blockd}[5]{\multirow{3}{*}{\(\left[\begin{array}{c}\text{3$\times$3, #2, stride #3}\\[-.1em] \text{3$\times$3, #2, stride #4} \end{array}\right]\)$\times$#5}
}

\begin{table}[t]
	\centering
	\caption{The network architecture of UNIQUE based on ResNet-34~\cite{he2016deep}. The nonlinearity and normalization are omitted for brevity}
	\begin{tabular}{l |c }
		\toprule
		Layer Name & Network Parameter\\
		\hline
		Convolution& {7$\times$7, 64, stride 2} \\
		\hline	
		Max Pooling & {3$\times$3, stride 2} \\
		\hline	
        \multirow{3}{*}{Residual Block 1}& \blockc{64}{1}{3}\\
        & \\
        &\\
		\hline
        \multirow{6}{*}{Residual Block 2}& \blockd{64}{128}{2}{1}{1}\\
        &\\
        &\\
        &\blockc{128}{1}{3}\\
        &\\
        &\\
        \hline
       \multirow{6}{*}{Residual Block 3}& \blockd{128}{256}{2}{1}{1}\\
        & \\
        &\\
        & \blockc{256}{1}{5}\\
        & \\
        &\\
        \hline
        \multirow{6}{*}{Residual Block 4}& \blockd{256}{512}{2}{1}{1}\\
        & \\
        &\\
        & \blockc{512}{1}{2}\\
        & \\
        &\\
        \hline
	   Bilinear Pooling	 & 0 \\
		 \hline
		 Fully Connected Layer&  262,144$ \times$2\\
		\bottomrule
	\end{tabular}
	\label{table:backbone}
\end{table}

\begin{table*}[t]
  \footnotesize
  \centering
  \caption{Median SRCC and PLCC results across ten sessions along with the average absolute deviation in the bracket. CLIVE stands for the LIVE Challenge Database. The databases used for training models are included in the bracket}\label{tab:overall}
  \begin{tabular}{l|cccccc}
      \toprule
    Database & {LIVE~\cite{sheikh2006statistical}} & {CSIQ~\cite{larson2010most}} & {KADID-10K~\cite{lin2019kadid}} & {BID~\cite{ciancio2011no}} &{CLIVE~\cite{ghadiyaram2016massive}} & {KonIQ-10K~\cite{hosu2020koniq}}\\
    \hline
    Criterion & \multicolumn{6}{c}{SRCC} \\
    \hline
       MS-SSIM~\cite{wang2003multiscale} & 0.951 ( {\color{grey}$\pm$ 0.007}) & 0.910 ({\color{grey}$\pm$  0.015}) & 0.821 ( {\color{grey}$\pm$ 0.011}) & -- & -- & -- \\
       NLPD~\cite{Laparra:17} & 0.942 ({\color{grey}$\pm$ 0.010}) & 0.937 ({\color{grey}$\pm$ 0.010})  & 0.822 ({\color{grey}$\pm$ 0.013})  & -- & -- & --  \\
       DISTS~\cite{ding2020iqa} & 0.955 ({\color{grey}$\pm$ 0.006}) & 0.944 ({\color{grey}$\pm$ 0.015}) & 0.892 ({\color{grey}$\pm$ 0.007}) & -- & -- & --  \\
     \hline
        NIQE~\cite{mittal2013making} & 0.906 ({\color{grey}$\pm$ 0.013}) & 0.632 ({\color{grey}$\pm$ 0.036}) & 0.374 ({\color{grey}$\pm$ 0.024}) & 0.468 ({\color{grey}$\pm$ 0.081}) & 0.464 ({\color{grey}$\pm$ 0.055}) & 0.521 ({\color{grey}$\pm$ 0.019}) \\
       ILNIQE~\cite{zhang2015feature} & 0.907 ({\color{grey}$\pm$ 0.013})  & 0.832 ({\color{grey}$\pm$ 0.030}) & 0.531 ({\color{grey}$\pm$ 0.025})  & 0.516 ({\color{grey}$\pm$ 0.092})  & 0.469 ({\color{grey}$\pm$ 0.047})  & 0.507 ({\color{grey}$\pm$ 0.018}) \\
       dipIQ~\cite{ma2017dipiq} & 0.940 ({\color{grey}$\pm$ 0.010}) & 0.511 ({\color{grey}$\pm$ 0.036}) & 0.304 ({\color{grey}$\pm$ 0.010}) & 0.009 ({\color{grey}$\pm$ 0.077})  & 0.187 ({\color{grey}$\pm$ 0.047}) & 0.228 ({\color{grey}$\pm$ 0.016})\\
        Ma19~\cite{ma2019blind} & 0.935 ({\color{grey}$\pm$ 0.060}) & 0.917 ({\color{grey}$\pm$ 0.043}) & 0.466 ({\color{grey}$\pm$ 0.062})  & 0.316 ({\color{grey}$\pm$ 0.091})  & 0.348 ({\color{grey}$\pm$ 0.048})  & 0.365 ({\color{grey}$\pm$ 0.042})\\
    \hline
      MEON (LIVE)~\cite{Ma2018End}  & -- & 0.726 ({\color{grey}$\pm$ 0.036}) & 0.234 ({\color{grey}$\pm$ 0.035})  & 0.100 ({\color{grey}$\pm$ 0.090})  & 0.378 ({\color{grey}$\pm$ 0.030}) & 0.145 ({\color{grey}$\pm$ 0.015})  \\
      deepIQA (LIVE)~\cite{bosse2016deep} & -- & 0.645 ({\color{grey}$\pm$ 0.029}) & 0.270 ({\color{grey}$\pm$ 0.041}) & -0.043 ({\color{grey}$\pm$ 0.105}) & 0.076 ({\color{grey}$\pm$ 0.059}) & -0.064 ({\color{grey}$\pm$ 0.027}) \\
      deepIQA (TID2013) & 0.834 ({\color{grey}$\pm$ 0.025}) & 0.679 ({\color{grey}$\pm$ 0.043}) & 0.559 ({\color{grey}$\pm$ 0.062}) & 0.236 ({\color{grey}$\pm$ 0.087}) & 0.048 ({\color{grey}$\pm$ 0.048}) & 0.182 ({\color{grey}$\pm$ 0.042})  \\
      RankIQA (LIVE)~\cite{liu2017rankiqa} & -- & 0.711 ({\color{grey}$\pm$ 0.011}) & 0.436 ({\color{grey}$\pm$ 0.025}) & 0.324 ({\color{grey}$\pm$ 0.062}) & 0.451 ({\color{grey}$\pm$ 0.047})& 0.617 ({\color{grey}$\pm$ 0.016}) \\
      RankIQA (TID2013) & 0.599 ({\color{grey}$\pm$ 0.026}) & 0.667 ({\color{grey}$\pm$ 0.079})  & 0.477 ({\color{grey}$\pm$ 0.028}) & 0.217 ({\color{grey}$\pm$ 0.088}) & 0.289 ({\color{grey}$\pm$ 0.051})& 0.460 ({\color{grey}$\pm$ 0.017}) \\
      PQR (BID)~\cite{zeng2018blind} & 0.663 ({\color{grey}$\pm$ 0.069}) & 0.522 ({\color{grey}$\pm$ 0.058}) & 0.321 ({\color{grey}$\pm$ 0.038}) & -- & 0.691 ({\color{grey}$\pm$ 0.026}) & 0.614 ({\color{grey}$\pm$ 0.014})\\
      PQR (KADID-10K) & 0.916 ({\color{grey}$\pm$ 0.016}) & 0.803 ({\color{grey}$\pm$ 0.086}) & -- & 0.358 ({\color{grey}$\pm$ 0.081})  & 0.439 ({\color{grey}$\pm$ 0.063}) & 0.485 ({\color{grey}$\pm$ 0.016}) \\
      PQR (KonIQ-10K) & 0.741 ({\color{grey}$\pm$ 0.058}) & 0.710 ({\color{grey}$\pm$ 0.045}) & 0.583 ({\color{grey}$\pm$ 0.025}) & 0.729 ({\color{grey}$\pm$ 0.045}) & 0.766 ({\color{grey}$\pm$ 0.026}) & -- \\
      PaQ-2-PiQ (LIVE Patch)~\cite{ying2020from} & 0.472 ({\color{grey}$\pm$ 0.048}) & 0.555 ({\color{grey}$\pm$ 0.055}) & 0.379 ({\color{grey}$\pm$ 0.026}) & 0.682 ({\color{grey}$\pm$ 0.038}) & 0.719 ({\color{grey}$\pm$ 0.027}) & 0.722 ({\color{grey}$\pm$ 0.009})\\
      DB-CNN (CSIQ)~\cite{zhang2020blind} & 0.855 ({\color{grey}$\pm$ 0.034}) & -- & 0.501 ({\color{grey}$\pm$ 0.025}) & 0.329 ({\color{grey}$\pm$ 0.092}) & 0.451 ({\color{grey}$\pm$ 0.056}) & 0.499 ({\color{grey}$\pm$ 0.016})\\
      DB-CNN (LIVE Challenge) & 0.723 ({\color{grey}$\pm$ 0.044}) & 0.691 ({\color{grey}$\pm$ 0.056}) & 0.488 ({\color{grey}$\pm$ 0.036}) & 0.809 ({\color{grey}$\pm$ 0.023}) & -- & 0.770 ({\color{grey}$\pm$ 0.010})\\
      KonCept512 (KonIQ-10K) & 0.776 ({\color{grey}$\pm$ 0.024}) & 0.639 ({\color{grey}$\pm$ 0.050}) & 0.491 ({\color{grey}$\pm$ 0.033}) & 0.800 ({\color{grey}$\pm$ 0.041}) & 0.781({\color{grey}$\pm$ 0.029}) & {\bf 0.917} ({\color{grey}$\pm$ 0.004})\\
     \hline
       UNIQUE (All databases)& {\bf 0.969} ({\color{grey}$\pm$ 0.008}) & {\bf 0.902} ({\color{grey}$\pm$ 0.045}) & {\bf 0.878} ({\color{grey}$\pm$ 0.016}) & {\bf 0.858} ({\color{grey}$\pm$ 0.032}) & {\bf 0.854} ({\color{grey}$\pm$ 0.019}) & 0.896 ({\color{grey}$\pm$ 0.006})\\
    \midrule
    Criterion  & \multicolumn{6}{c}{PLCC}\\
    \hline
       MS-SSIM~\cite{wang2003multiscale} & 0.941 ({\color{grey}$\pm$ 0.006}) & 0.897 ({\color{grey}$\pm$ 0.014}) & 0.819 ({\color{grey}$\pm$ 0.011}) & -- & -- & --  \\
       NLPD~\cite{Laparra:17} & 0.937 ({\color{grey}$\pm$ 0.008}) & 0.930 ({\color{grey}$\pm$ 0.015}) & 0.821 ({\color{grey}$\pm$ 0.014}) & -- & -- & --  \\
       DISTS~\cite{ding2020iqa} & 0.955 ({\color{grey}$\pm$ 0.006}) & 0.946 ({\color{grey}$\pm$ 0.015}) & 0.892 ({\color{grey}$\pm$ 0.007})& -- & -- & --  \\
     \hline
        NIQE~\cite{mittal2013making} & 0.908 ({\color{grey}$\pm$ 0.013}) & 0.726 ({\color{grey}$\pm$ 0.034}) & 0.428 ({\color{grey}$\pm$ 0.021}) & 0.461 ({\color{grey}$\pm$ 0.073}) & 0.515 ({\color{grey}$\pm$ 0.050}) & 0.529 ({\color{grey}$\pm$ 0.014})\\
       ILNIQE~\cite{zhang2015feature} & 0.912 ({\color{grey}$\pm$ 0.012}) & 0.873 ({\color{grey}$\pm$ 0.026}) & 0.573 ({\color{grey}$\pm$ 0.027})& 0.533 ({\color{grey}$\pm$ 0.094}) & 0.536 ({\color{grey}$\pm$ 0.030}) &0.534 ({\color{grey}$\pm$ 0.018})\\
       dipIQ~\cite{ma2017dipiq}  & 0.933 ({\color{grey}$\pm$ 0.011}) & 0.778 ({\color{grey}$\pm$ 0.020}) & 0.402 ({\color{grey}$\pm$ 0.011}) & 0.346 ({\color{grey}$\pm$ 0.069}) & 0.290 ({\color{grey}$\pm$ 0.042}) & 0.437 ({\color{grey}$\pm$ 0.009}) \\
        Ma19~\cite{ma2019blind} & 0.934 ({\color{grey}$\pm$ 0.063}) & 0.926 ({\color{grey}$\pm$ 0.024}) & 0.500 ({\color{grey}$\pm$ 0.061}) & 0.348 ({\color{grey}$\pm$ 0.057}) & 0.400 ({\color{grey}$\pm$ 0.047})& 0.416 ({\color{grey}$\pm$ 0.042}) \\
    \hline
      MEON (LIVE)~\cite{Ma2018End}  & -- & 0.787 ({\color{grey}$\pm$ 0.020}) & 0.410 ({\color{grey}$\pm$ 0.007}) & 0.217 ({\color{grey}$\pm$ 0.057}) & 0.477 ({\color{grey}$\pm$ 0.028}) & 0.242 ({\color{grey}$\pm$ 0.014}) \\
      deepIQA (LIVE)~\cite{bosse2016deep} & -- & 0.730 ({\color{grey}$\pm$ 0.029}) & 0.309 ({\color{grey}$\pm$ 0.059}) & 0.127 ({\color{grey}$\pm$ 0.083}) & 0.162 ({\color{grey}$\pm$ 0.071}) & 0.088 ({\color{grey}$\pm$ 0.027}) \\
      deepIQA (TID2013) & 0.853 ({\color{grey}$\pm$ 0.013}) & 0.771 ({\color{grey}$\pm$ 0.024}) & 0.573 ({\color{grey}$\pm$ 0.061}) & 0.282 ({\color{grey}$\pm$ 0.067}) & 0.150 ({\color{grey}$\pm$ 0.038}) & 0.272 ({\color{grey}$\pm$ 0.042}) \\
      RankIQA (LIVE)~\cite{liu2017rankiqa} & -- & 0.790 ({\color{grey}$\pm$ 0.008}) & 0.488 ({\color{grey}$\pm$ 0.017}) & 0.350 ({\color{grey}$\pm$ 0.052}) & 0.503 ({\color{grey}$\pm$ 0.043})& 0.631 ({\color{grey}$\pm$ 0.014})\\
      RankIQA (TID2013) & 0.633 ({\color{grey}$\pm$ 0.030}) & 0.778 ({\color{grey}$\pm$ 0.093}) & 0.504 ({\color{grey}$\pm$ 0.031}) & 0.306 ({\color{grey}$\pm$ 0.071}) & 0.313 ({\color{grey}$\pm$ 0.051}) & 0.471 ({\color{grey}$\pm$ 0.015}) \\
      PQR (BID)~\cite{zeng2018blind} & 0.673 ({\color{grey}$\pm$ 0.056}) & 0.612 ({\color{grey}$\pm$ 0.032})& 0.403 ({\color{grey}$\pm$ 0.030}) & -- & 0.740 ({\color{grey}$\pm$ 0.014}) & 0.650 ({\color{grey}$\pm$ 0.014}) \\
      PQR (KADID-10K) & 0.927 ({\color{grey}$\pm$ 0.013}) & 0.870 ({\color{grey}$\pm$ 0.051}) & -- & 0.405 ({\color{grey}$\pm$ 0.063}) & 0.501 ({\color{grey}$\pm$ 0.051}) & 0.484 ({\color{grey}$\pm$ 0.018}) \\
      PQR (KonIQ-10K) & 0.760 ({\color{grey}$\pm$ 0.060}) & 0.737 ({\color{grey}$\pm$ 0.042})& 0.604 ({\color{grey}$\pm$ 0.025}) & 0.739 ({\color{grey}$\pm$ 0.039}) & 0.826 ({\color{grey}$\pm$ 0.021}) & -- \\
      PaQ-2-PiQ (LIVE Patch)~\cite{ying2020from} & 0.559 ({\color{grey}$\pm$ 0.037}) & 0.658 ({\color{grey}$\pm$ 0.034}) & 0.429 ({\color{grey}$\pm$ 0.026}) & 0.713 ({\color{grey}$\pm$ 0.030}) & 0.778 ({\color{grey}$\pm$ 0.014}) & 0.735 ({\color{grey}$\pm$ 0.010})\\
      DB-CNN (CSIQ)~\cite{zhang2020blind} & 0.854 ({\color{grey}$\pm$ 0.033})& -- & 0.569 ({\color{grey}$\pm$ 0.023}) & 0.383  ({\color{grey}$\pm$ 0.088})& 0.472 ({\color{grey}$\pm$ 0.043})& 0.515 ({\color{grey}$\pm$ 0.015})\\
      DB-CNN (LIVE Challenge) & 0.754 ({\color{grey}$\pm$ 0.049}) & 0.685 ({\color{grey}$\pm$ 0.050}) & 0.529 ({\color{grey}$\pm$ 0.031})& 0.832 ({\color{grey}$\pm$ 0.024}) & -- & 0.825 ({\color{grey}$\pm$ 0.007}) \\
      KonCept512 (KonIQ-10K) & 0.772 ({\color{grey}$\pm$ 0.019}) & 0.663 ({\color{grey}$\pm$ 0.029}) & 0.508 ({\color{grey}$\pm$ 0.021}) & 0.818 ({\color{grey}$\pm$ 0.031}) & 0.844({\color{grey}$\pm$ 0.023}) & {\bf 0.931} ({\color{grey}$\pm$ 0.003})\\
     \hline
       UNIQUE (All databases) & {\bf 0.968} ({\color{grey}$\pm$ 0.004})  & {\bf 0.927} ({\color{grey}$\pm$ 0.032})& {\bf 0.876} ({\color{grey}$\pm$ 0.016}) & {\bf 0.873} ({\color{grey}$\pm$ 0.025}) & {\bf 0.890} ({\color{grey}$\pm$ 0.011}) & 0.901 ({\color{grey}$\pm$ 0.006})\\
     \bottomrule
   \end{tabular}
\end{table*}

\section{Experiments}\label{sec:exp}
In this section, we first present the experimental setups, including the IQA databases, the evaluation protocols, and the training details of UNIQUE. We then compare UNIQUE with several state-of-the-art BIQA models on existing IQA databases as well as the gMAD competition~\cite{ma2020group}. Moreover, we verify the promise of the proposed training strategy by comparing it with alternative training schemes, testing it in a cross-database setting, and using it to improve existing BIQA models.

\subsection{Experimental Setups}\label{subsec:exp_setup}
We chose six IQA databases (summarized in  Table~\ref{tab:database}), among which LIVE~\cite{sheikh2006statistical}, CSIQ~\cite{larson2010most}, and KADID-10K~\cite{lin2019kadid} contain synthetic distortions, while LIVE Challenge~\cite{ghadiyaram2016massive}, BID~\cite{ciancio2011no}, and KonIQ-10K~\cite{hosu2020koniq} include realistic distortions. All selected databases provide the stds of subjective quality scores (see Fig.~\ref{fig:gt_curve}).  We excluded TID2013~\cite{ponomarenko2013color} in our experiments because the MOS is computed by the number of winning times in a maximum of nine paired comparisons without suitable psychometric scaling~\cite{mikhailiuk2018psychometric}, and does not satisfy the Gaussian assumption. We refer the interested readers to our preliminary work~\cite{zhang2020learning} for the results on TID2013~\cite{ponomarenko2013color}.

We randomly sampled $80\%$ images from each database to construct the training set and leave the remaining $20\%$ for testing. Regarding synthetic databases LIVE, CSIQ, and KADID-10K, we split the training and test sets according to the reference images in order to ensure content independence. We adopted two performance criteria: Spearman rank-order correlation coefficient (SRCC) and Pearson linear correlation coefficient (PLCC), which measure prediction monotonicity and precision~\cite{gao2020extended}, respectively. To reduce the bias caused by the randomness in training and test set splitting, we repeated this procedure ten times, and reported median SRCC and PLCC results.

We trained UNIQUE on $270,000$ image pairs by minimizing the objective in Eq. \eqref{eq:loss1} using Adam~\cite{Kingma2014adam}. The margin of the hinge loss $\xi$ and the trade-off parameter $\lambda$ were set to $0.025$ and $1$, respectively. Empirically, we found that the performance is insensitive to the two hyperparameters. We used a softplus function to constrain the predicted std $\sigma_{w}(x)$ to be positive. The parameters of UNIQUE based on  ResNet-34~\cite{he2016deep} were initialized with the weights pre-trained  on ImageNet~\cite{deng2009imagenet}. The parameters of the last fully connected layer were initialized by He's method  \cite{he2015delving}.  We set the initial learning rate to $10^{-4}$ with a decay factor of $10$ for every three epochs, and we trained UNIQUE twelve epochs. A warm-up training strategy was adopted: only the last fully connected layer was trained in the first three epochs with a mini-batch of $128$; for the remaining epochs, we fine-tuned the entire method with a mini-batch of $32$. During training, we re-scaled and cropped the images to $384 \times 384 \times 3$, keeping their aspect ratios. In all experiments, we tested on images of original size. We implemented UNIQUE using PyTorch, and made the  code publicly available at \url{https://github.com/zwx8981/UNIQUE}.

\subsection{Main Results}\label{subsec:exp results}

\begin{figure*}[t]
  \centering
  \includegraphics[width=1\textwidth]{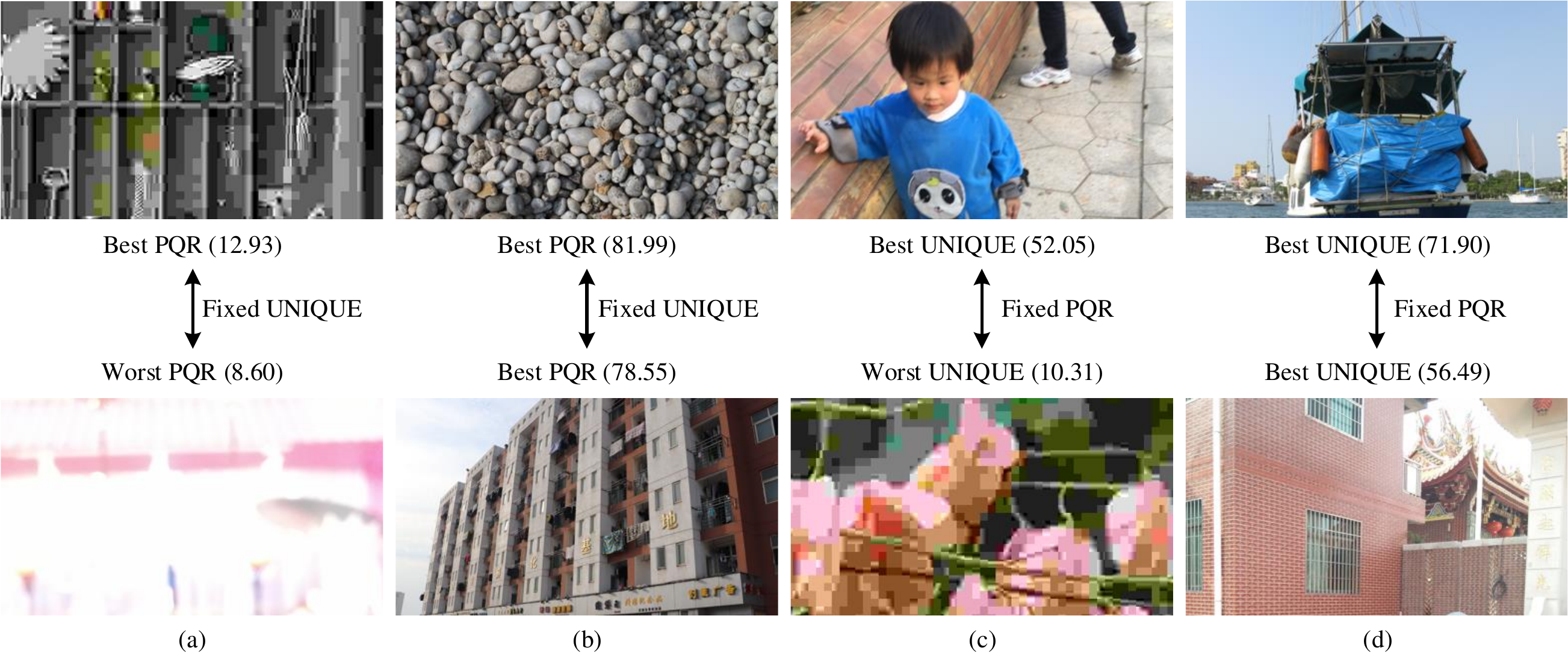}
  \caption{gMAD competition results between PQR~\cite{zeng2018blind} and UNIQUE.  The MOS of each image is shown in the bracket. (a) Fixed UNIQUE at the low-quality level. (b) Fixed UNIQUE at the high-quality level. (c) Fixed PQR at the low-quality level. (d) Fixed PQR at the high-quality level.}\label{fig:gmad1}
\end{figure*}

\begin{figure*}[t]
  \centering
  \includegraphics[width=1\textwidth]{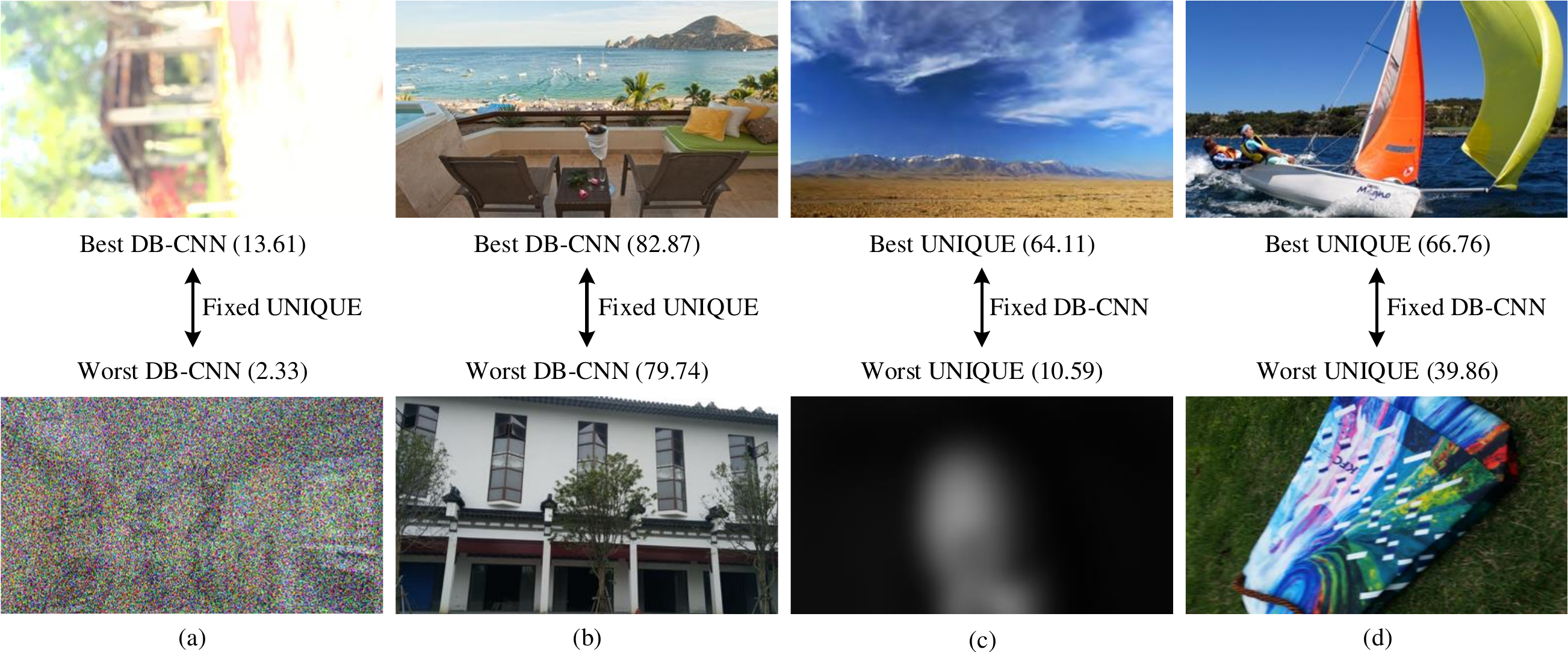}
  \caption{gMAD competition results between DB-CNN~\cite{zhang2020blind} and UNIQUE. The MOS of each image is shown in the bracket. (a) Fixed UNIQUE at the low-quality level. (b) Fixed UNIQUE at the high-quality level. (c) Fixed DB-CNN at the low-quality level. (d) Fixed DB-CNN at the high-quality level. }\label{fig:gmad2}
\end{figure*}

\begin{figure*}[t]
    \centering
    \captionsetup{justification=centering}
    \subfloat[LIVE]{\includegraphics[width=0.33\textwidth]{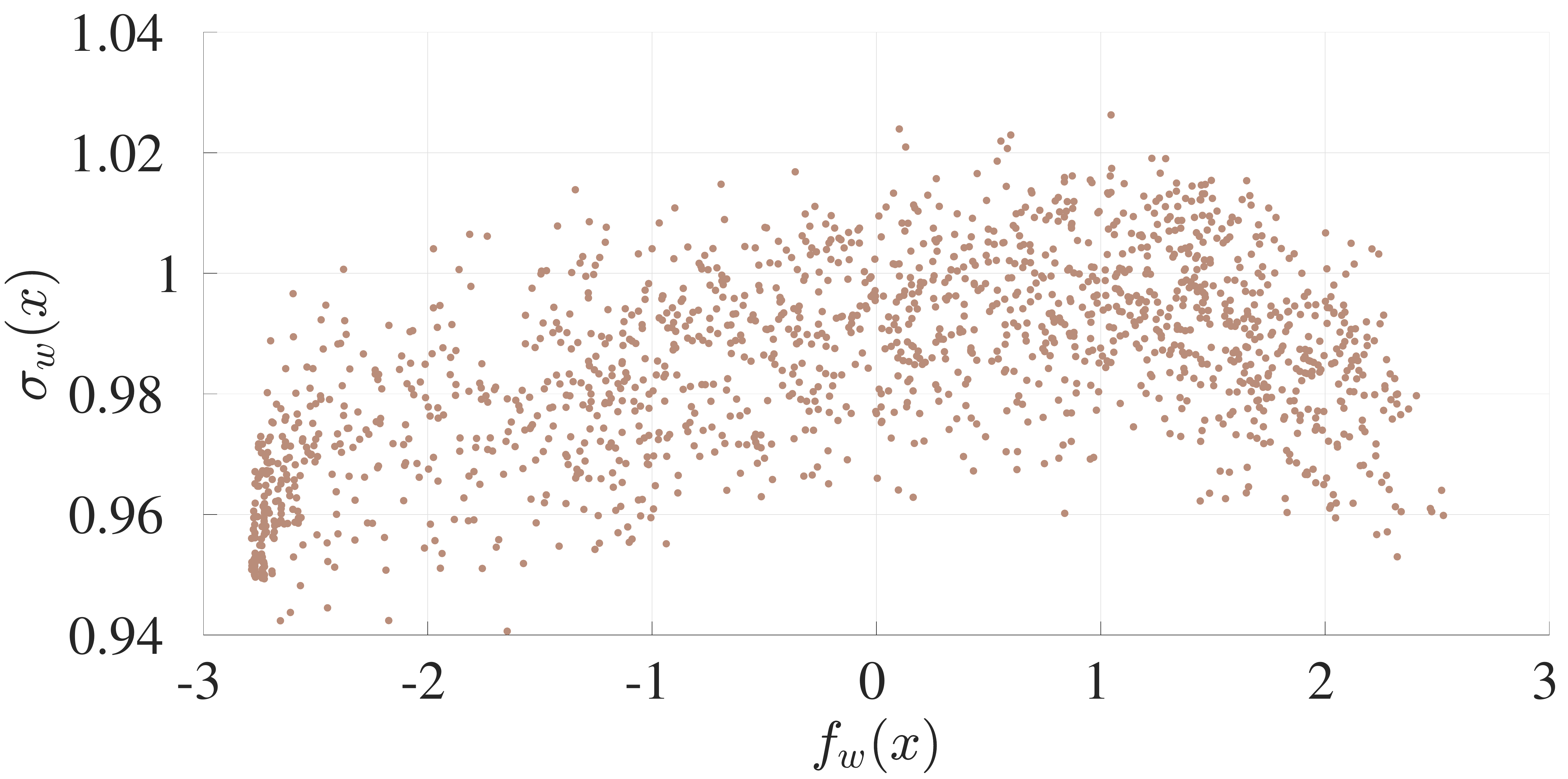}}
    \subfloat[CSIQ]{\includegraphics[width=0.33\textwidth]{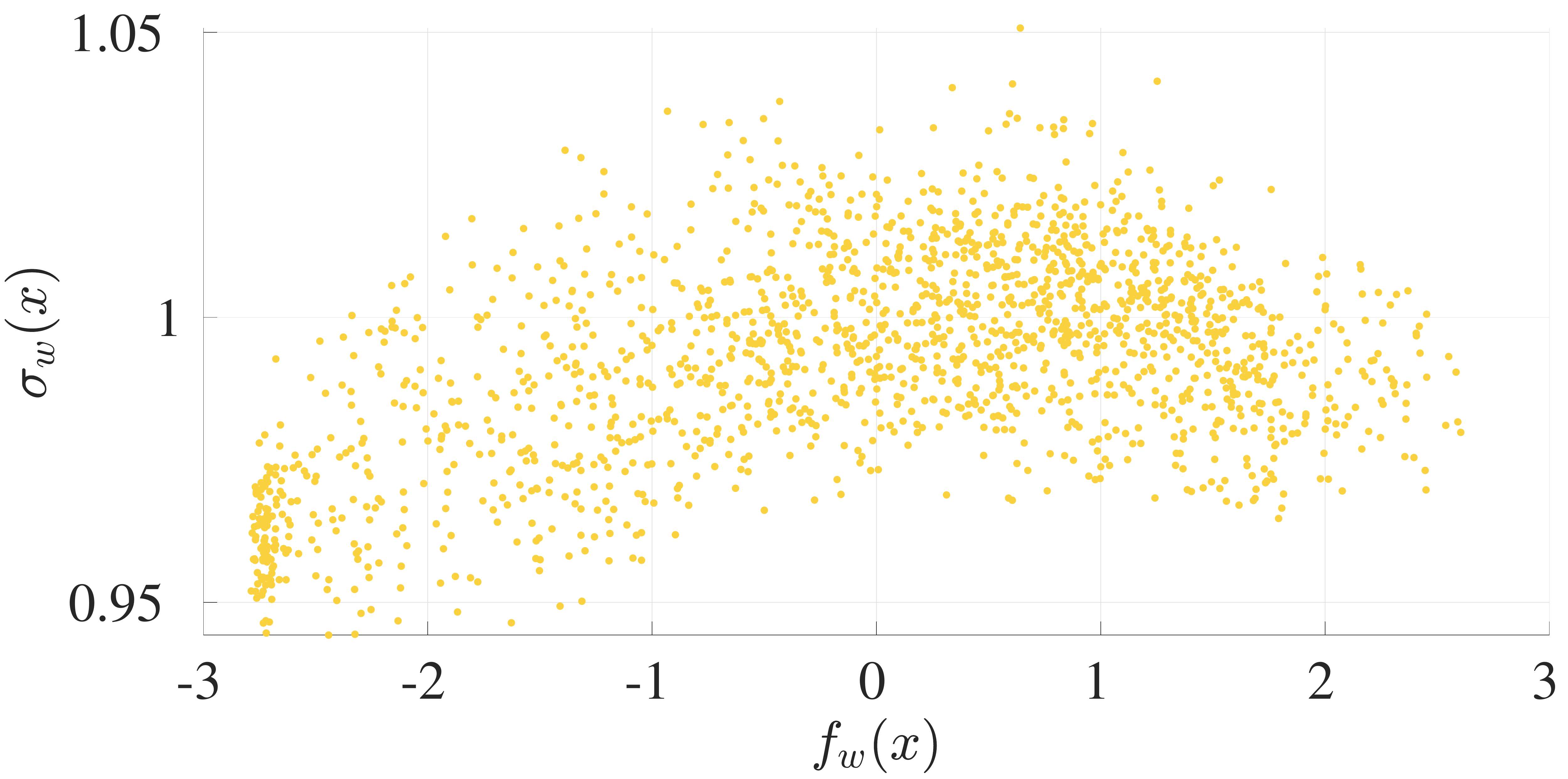}}
    \subfloat[KADID-10K]{\includegraphics[width=0.33\textwidth]{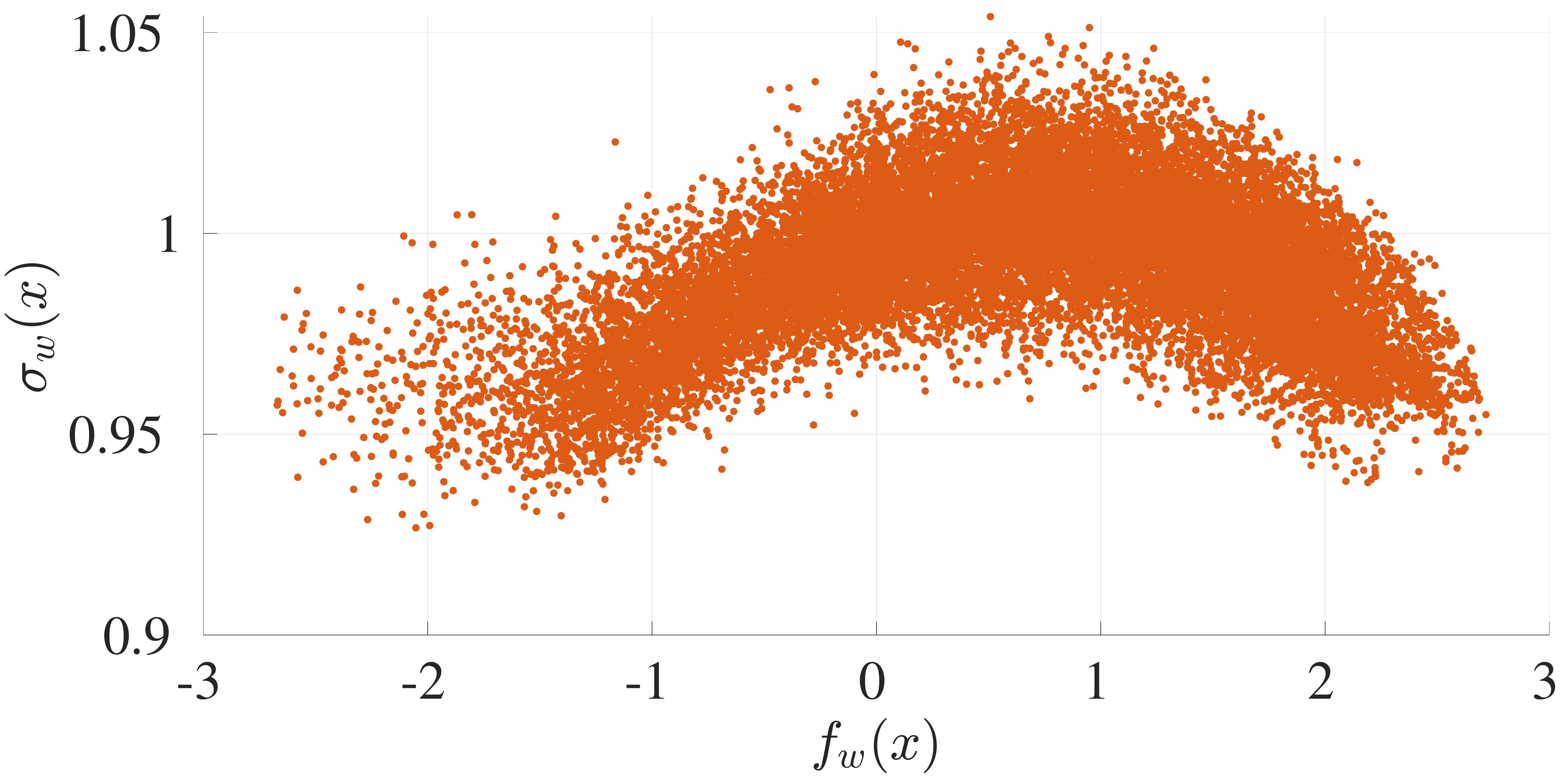}}
    \vspace{0em}
    \subfloat[BID]{\includegraphics[width=0.33\textwidth]{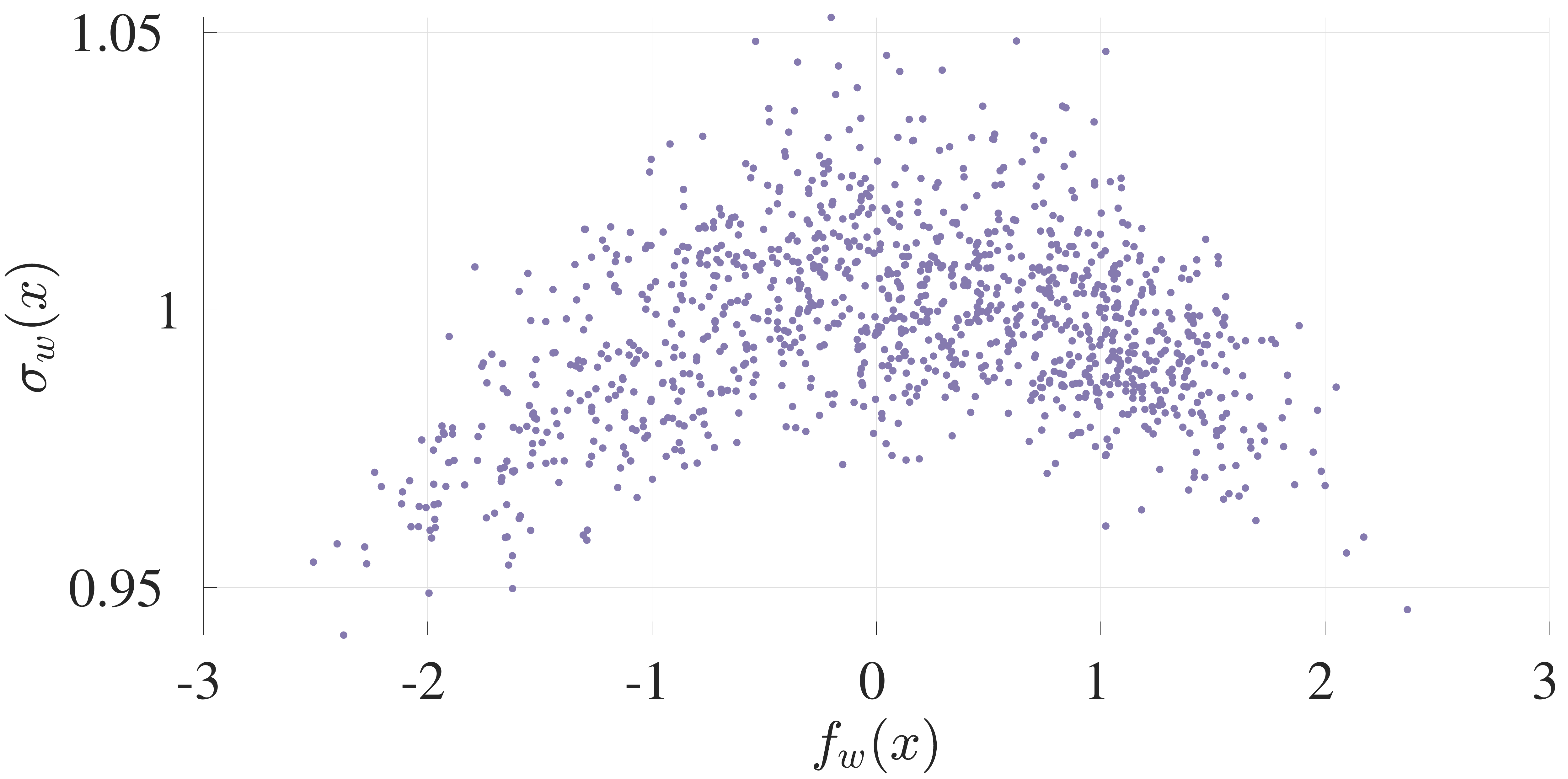}}
    \subfloat[LIVE Challenge]{\includegraphics[width=0.33\textwidth]{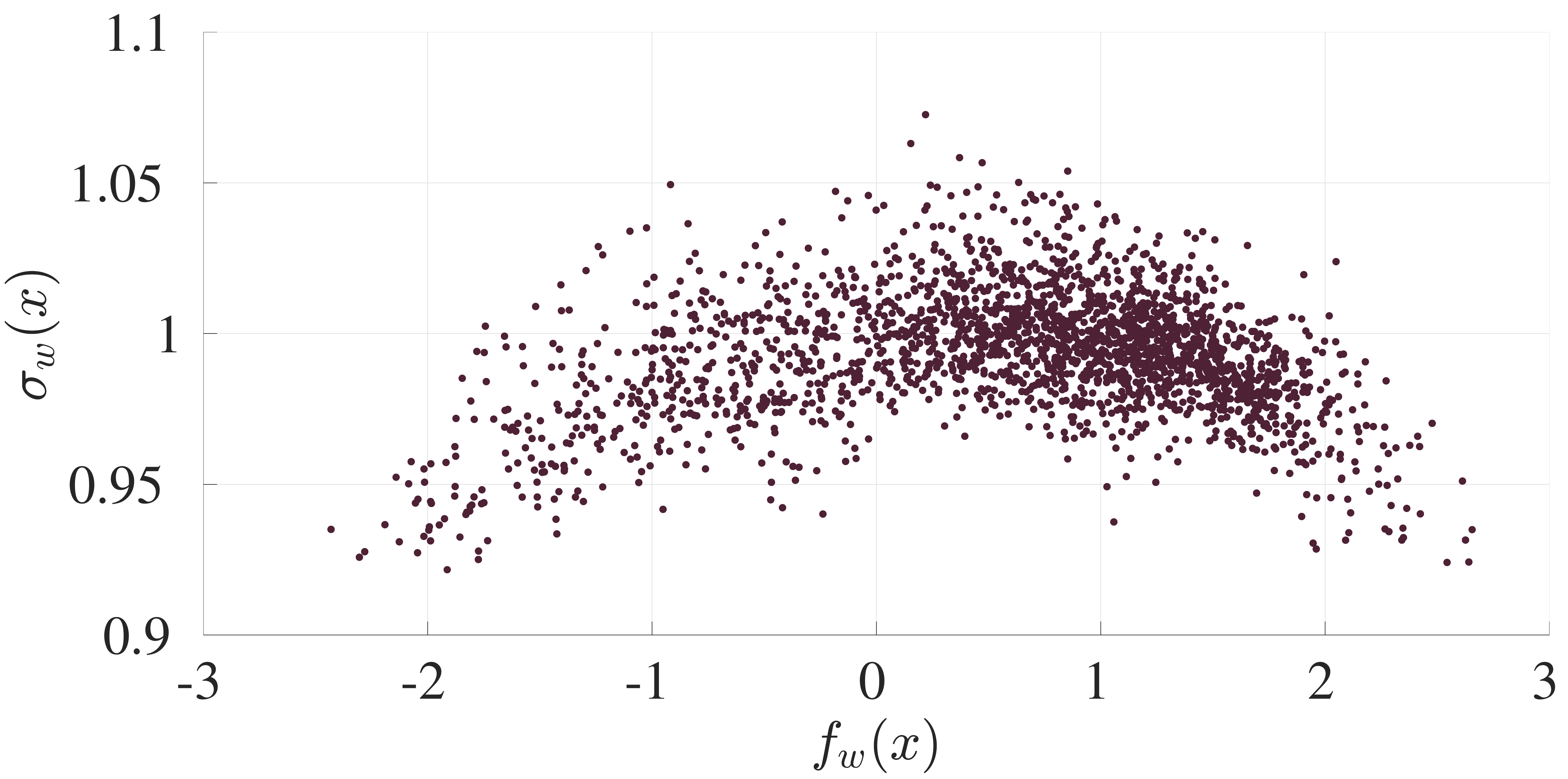}}
    \subfloat[KonIQ-10K]{\includegraphics[width=0.33\textwidth]{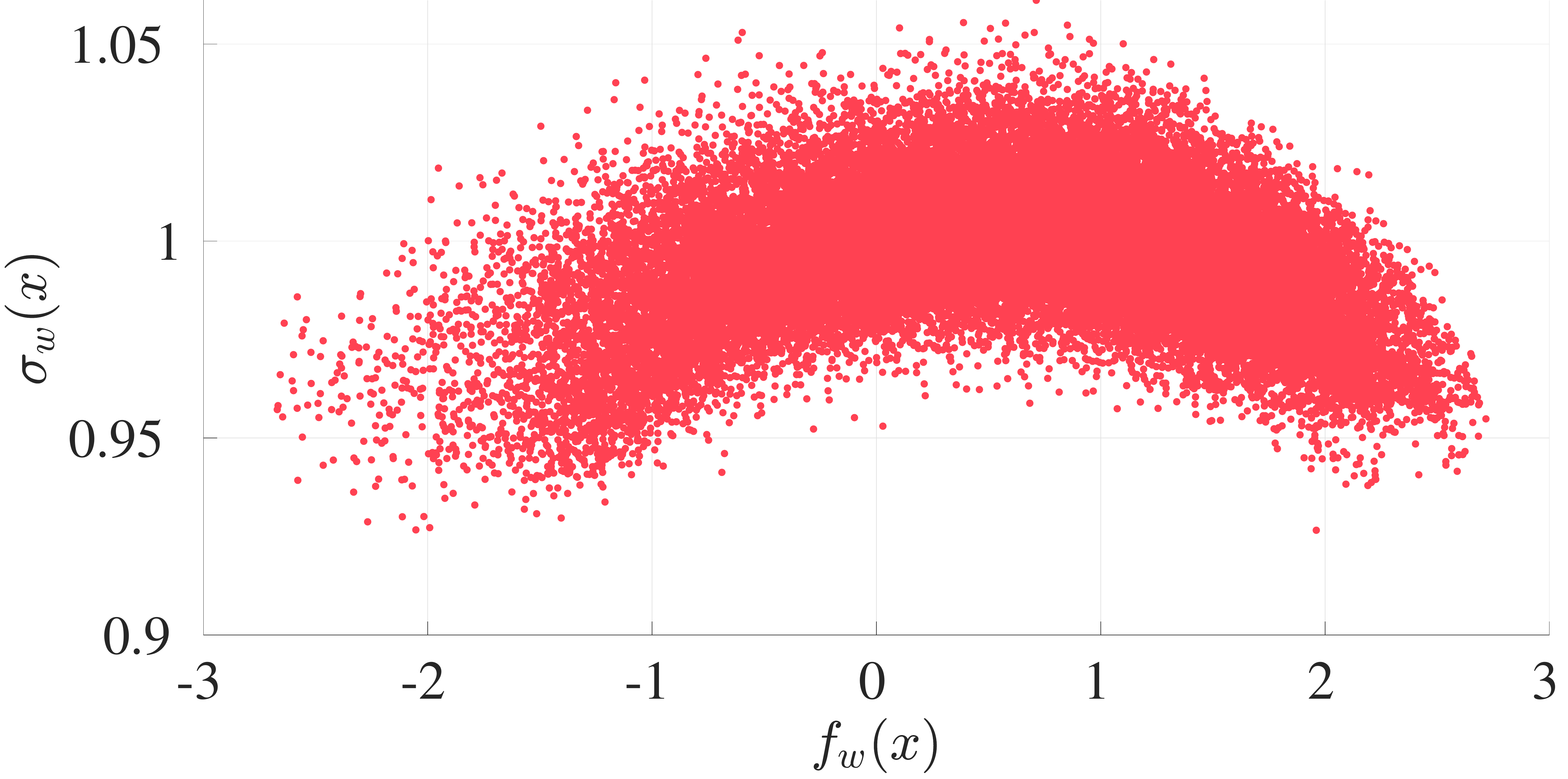}}
    %\vspace{-.2cm}
  \caption{Scatter plots of means against stds of images from six IQA databases predicted by UNIQUE with the hinge loss. $f_{w}(x)$ and $\sigma_{w}(x)$ indicate the predicted mean and std of image $x$, respectively.}
\label{fig:all_curves1}
\end{figure*}

\begin{figure*}[t]
    \centering
    \captionsetup{justification=centering}
    \subfloat[LIVE]{\includegraphics[width=0.33\textwidth]{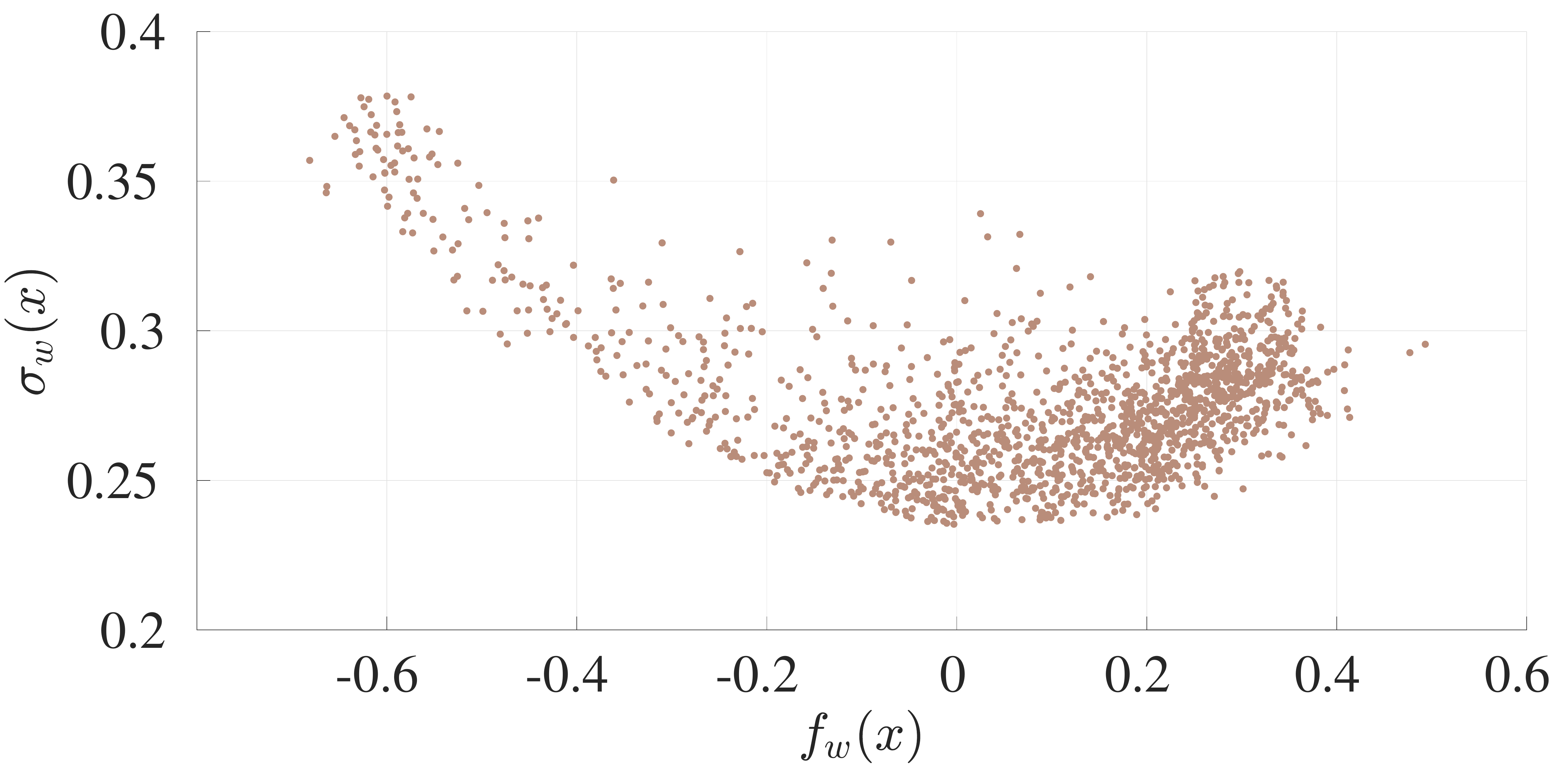}}
    \subfloat[CSIQ]{\includegraphics[width=0.33\textwidth]{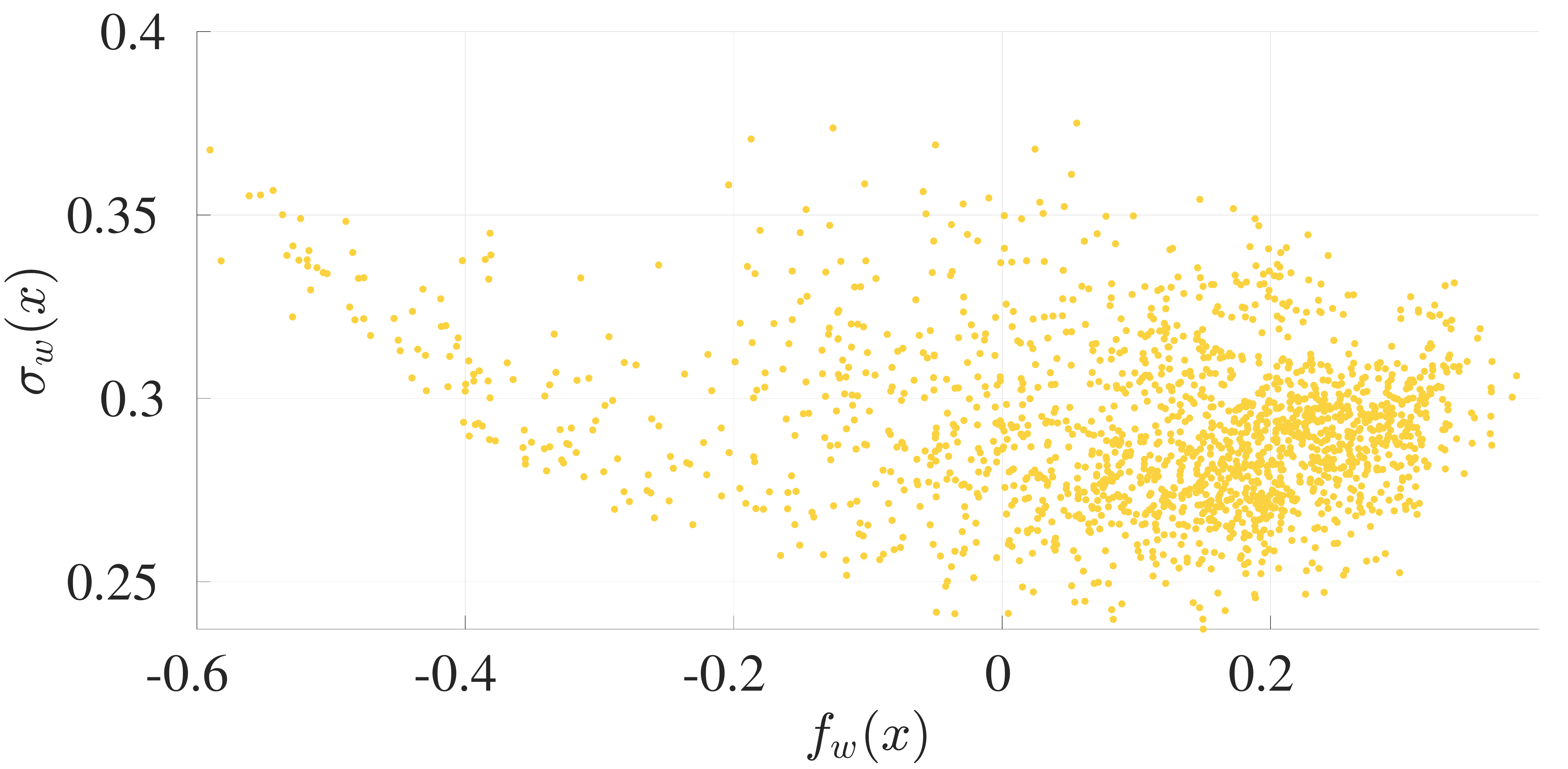}}
    \subfloat[KADID-10K]{\includegraphics[width=0.33\textwidth]{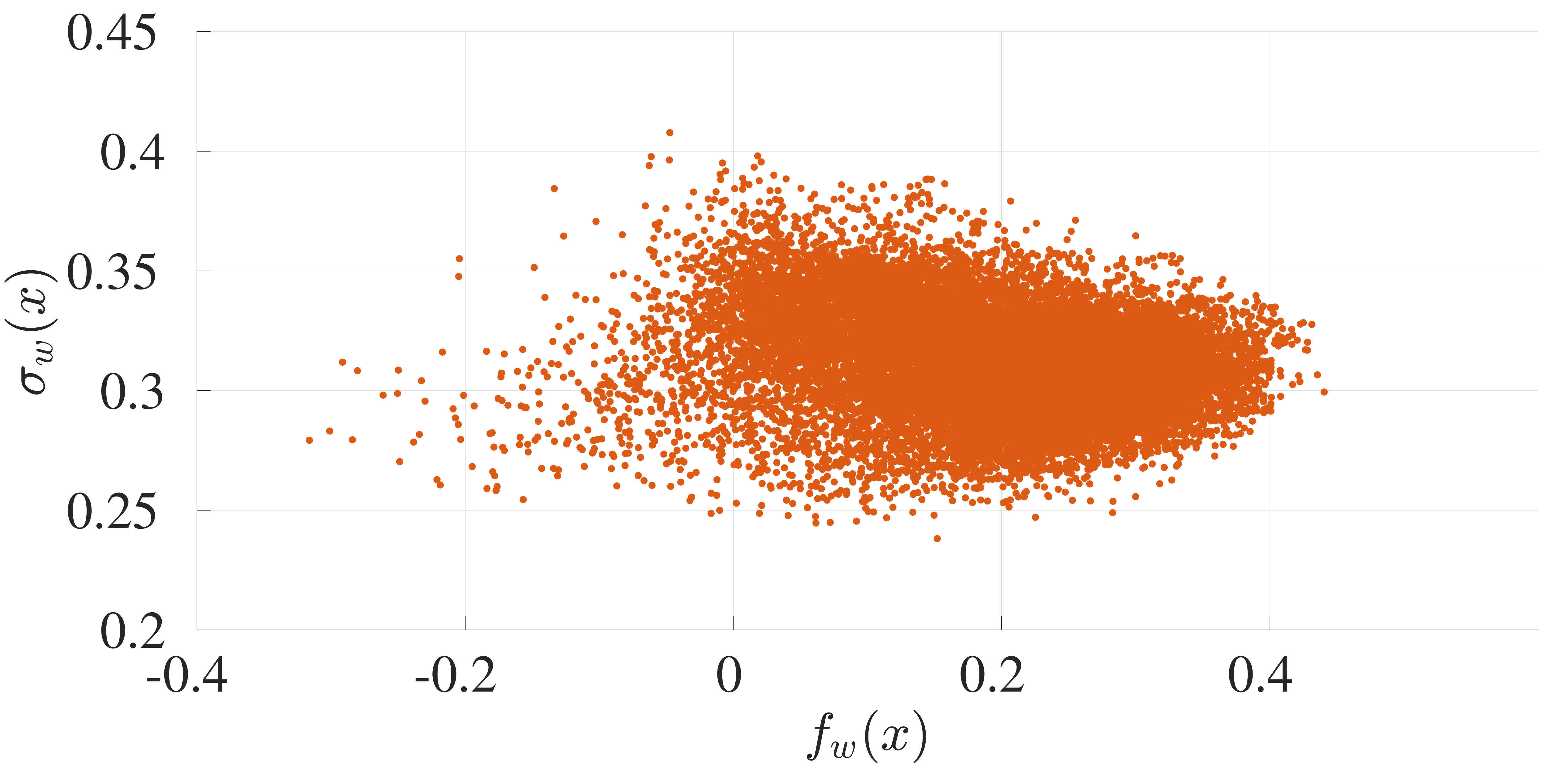}}
    \vspace{0em}
    \subfloat[BID]{\includegraphics[width=0.33\textwidth]{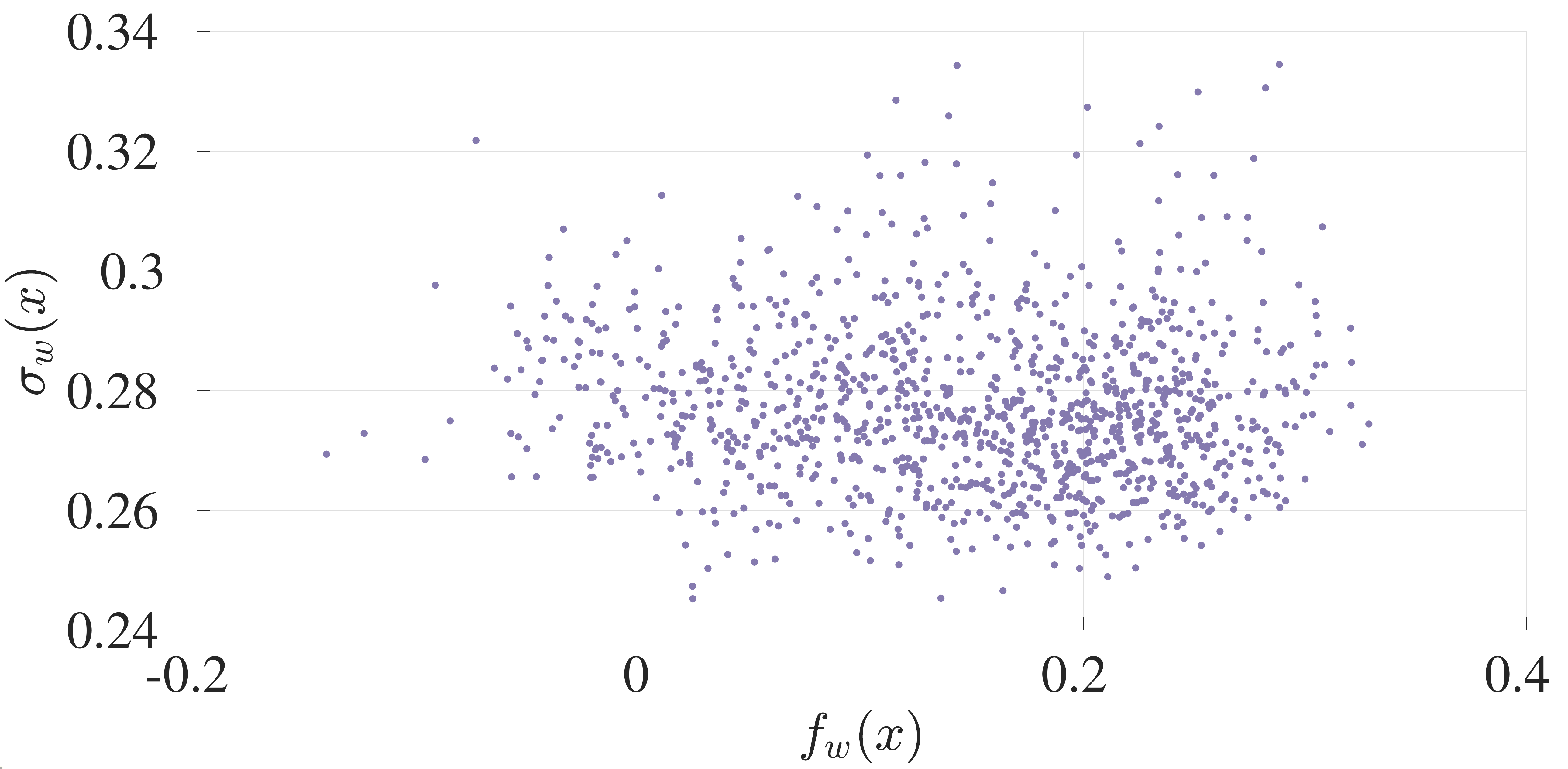}}
    \subfloat[LIVE Challenge]{\includegraphics[width=0.33\textwidth]{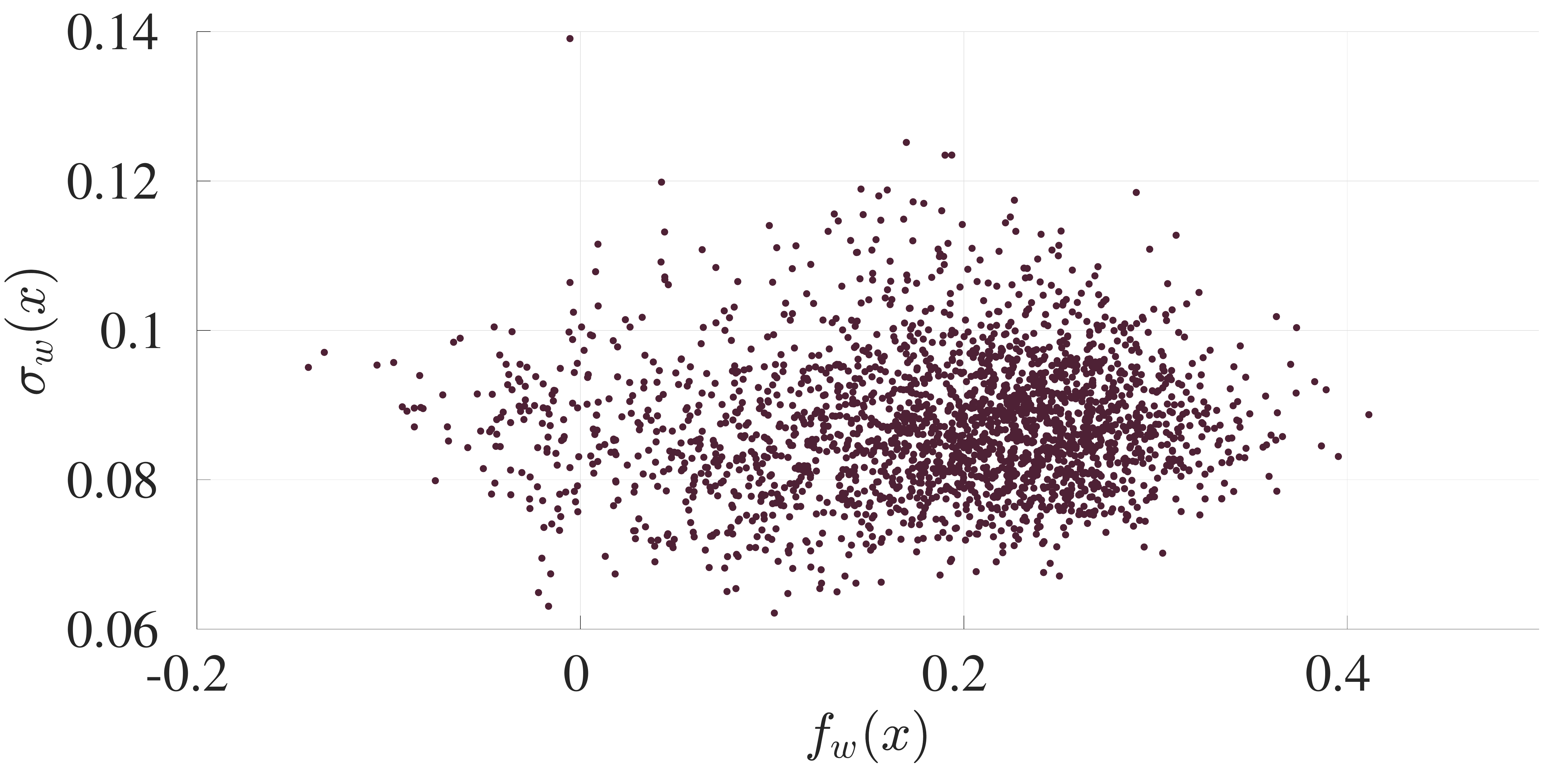}}
    \subfloat[KonIQ-10K]{\includegraphics[width=0.33\textwidth]{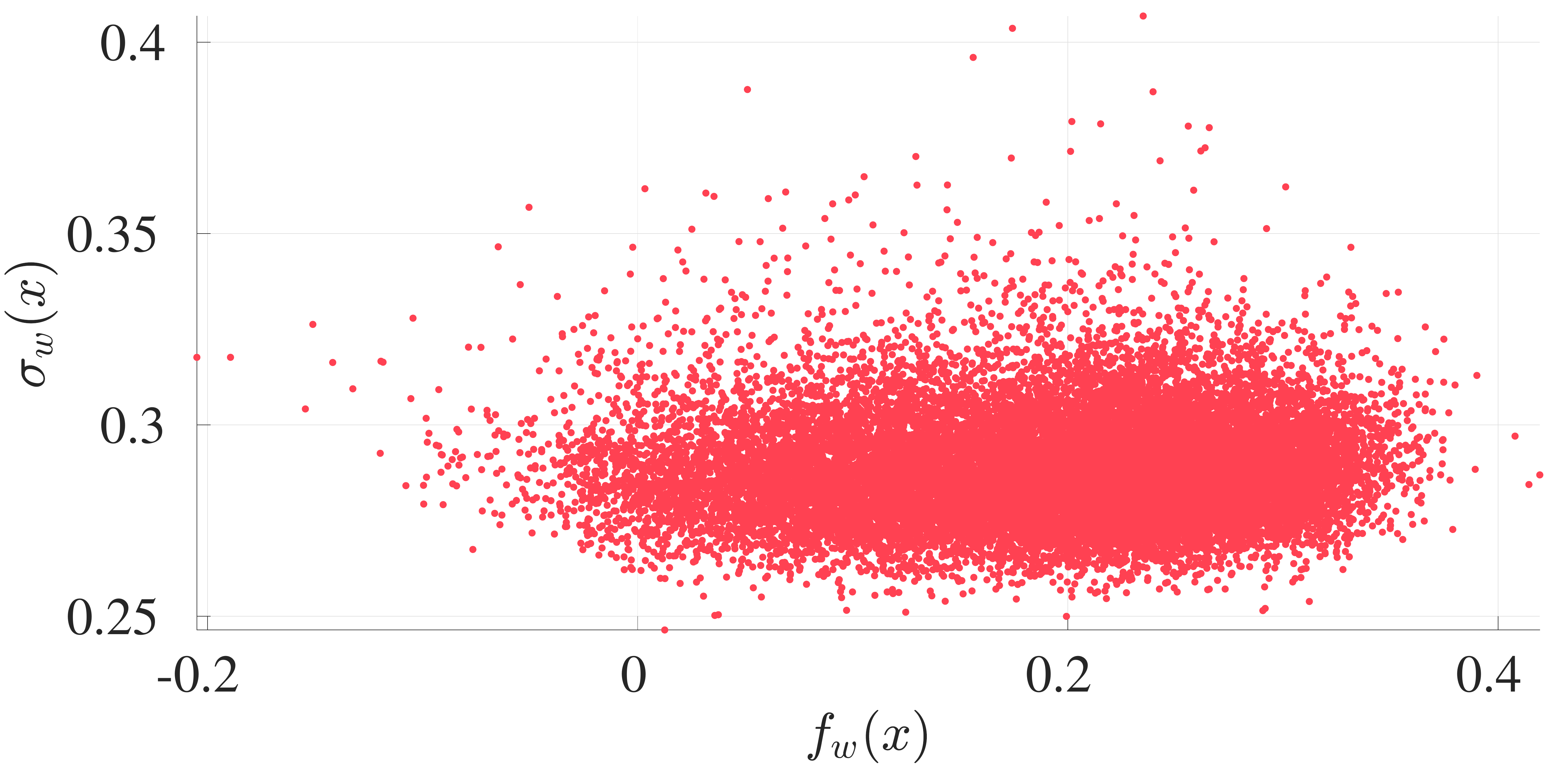}}
    %\vspace{-.2cm}
  \caption{Scatter plots of means against stds of images from six IQA databases predicted by UNIQUE without the hinge loss.}
\label{fig:all_curves2}
\end{figure*}

\subsubsection{Correlation Results}\label{subsubsec:trainstrat}
We compared the performance of UNIQUE against three full-reference IQA measures - MS-SSIM~\cite{wang2003multiscale}, NLPD~\cite{Laparra:17} and DISTS~\cite{ding2020iqa}, and ten BIQA models, including four knowledge-driven methods that do not require MOSs for training - NIQE~\cite{mittal2013making}, IL-NIQE~\cite{zhang2015feature}, dipIQ~\cite{ma2017dipiq} and Ma19~\cite{ma2019blind}, and six data-driven DNN-based methods - MEON~\cite{Ma2018End}, deepIQA~\cite{bosse2016deep}, RankIQA~\cite{liu2017rankiqa}, PQR~\cite{zeng2018blind}, PaQ-2-PiQ~\cite{ying2020from} and DB-CNN~\cite{zhang2020blind}. For the competing models, we either used the publicly available implementations or re-trained them on the specific databases using the training codes provided by the corresponding authors~\cite{zeng2018blind, zhang2020blind}. \textcolor{black}{The median SRCC and PLCC results along with the average absolute deviations across ten sessions are listed in Table~\ref{tab:overall}. Note that some methods have been trained on all images from one of the six databases. We thus excluded the corresponding results.} NIQE and its improved version ILNIQE do not perform well on realistic distortions and challenging synthetic distortions in KADID-10K~\cite{lin2019kadid}, despite the original goal of handling arbitrary distortions. dipIQ \cite{ma2017dipiq} and Ma19 \cite{ma2019blind} may only be able to deal with distortions included in the training set. This highlights the difficulty of distortion-aware BIQA methods to handle unseen distortions.

\begin{figure*}[t]
    \centering
    \captionsetup{justification=centering}
    \subfloat[LIVE Challenge,  $f_{w}(x)$ = 2.641]{\includegraphics[width=0.248\textwidth]{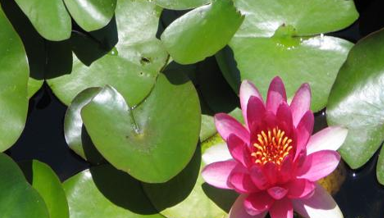}}\hskip.1em
    \subfloat[KADID-10K,  $f_{w}(x)$ = 2.626]{\includegraphics[width=0.248\textwidth]{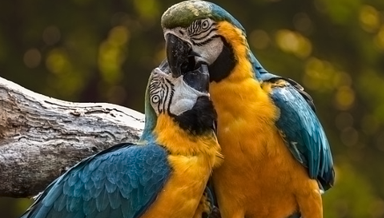}}\hskip.1em
    %\subfloat[pmos=2.5278 pstd:0.9598 Database: LIVE]{\includegraphics[width=0.248\textwidth]{figs/h_rank4}}\hskip.1em
    \subfloat[KonIQ-10K,  $f_{w}(x)$ = 2.520]{\includegraphics[width=0.248\textwidth]{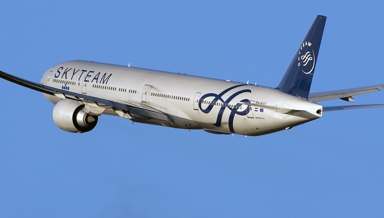}}\hskip.1em
    \subfloat[BID,  $f_{w}(x)$ = 1.548]{\includegraphics[width=0.248\textwidth]{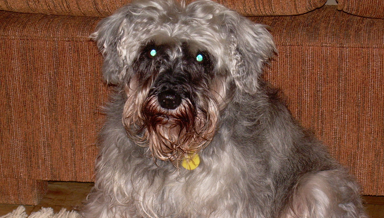}}
    \vspace{0em}
    \subfloat[KonIQ-10K,  $f_{w}(x)$ = 0.772]{\includegraphics[width=0.248\textwidth]{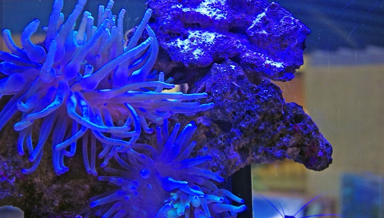}}\hskip.1em
    \subfloat[CSIQ,  $f_{w}(x)$ = 0.417]{\includegraphics[width=0.248\textwidth]{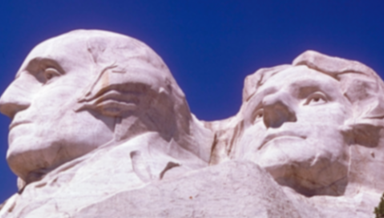}}\hskip.1em
    \subfloat[LIVE,  $f_{w}(x)$ = 0.264]{\includegraphics[width=0.248\textwidth]{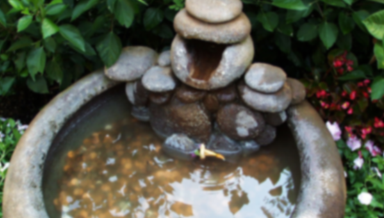}}\hskip.1em
    \subfloat[BID, $f_{w}(x)$ = 0.107]{\includegraphics[width=0.248\textwidth]{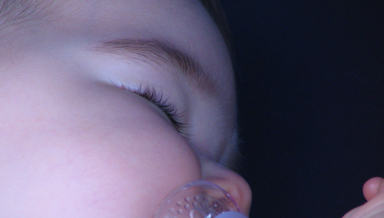}}
    \vspace{0em}
    \subfloat[LIVE Challenge,  $f_{w}(x)$ = -2.055]{\includegraphics[width=0.248\textwidth]{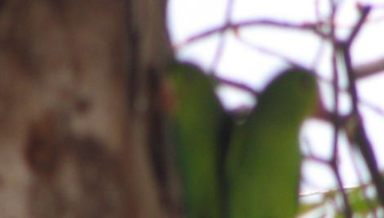}}\hskip.1em
    \subfloat[KADID-10K,  $f_{w}(x)$ = -2.388]{\includegraphics[width=0.248\textwidth]{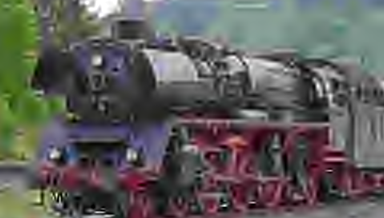}}\hskip.1em
    \subfloat[LIVE,  $f_{w}(x)$ = -2.785]{\includegraphics[width=0.248\textwidth]{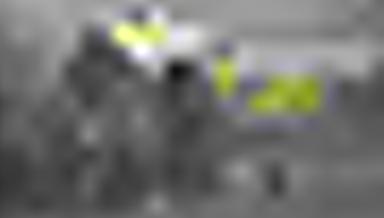}}\hskip.1em
    \subfloat[CSIQ,  $f_{w}(x)$ = -2.787]{\includegraphics[width=0.248\textwidth]{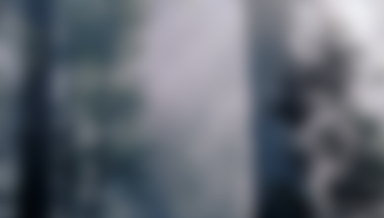}}
  \caption{Visual examples from different databases aligned in the learned perceptual scale. (a)-(d) Images with good predicted quality. (e)-(h) Images with fair predicted quality. (i)-(l) Images with poor predicted quality. In each row, images are arranged from left to right in descending order of predicted quality.}
\label{fig:qualitative}
\end{figure*}

We observe similar phenomena for deep learning-based methods
when facing the cross-distortion-scenario challenge. Despite large-scale pre-training, MEON fine-tuned on LIVE~\cite{sheikh2006statistical} does not generalize to other databases with different distortion types and scenarios. Being exposed to more synthetic distortion types in TID2013~\cite{ponomarenko2013color}, deepIQA and RankIQA achieve better performance on KADID-10K~\cite{lin2019kadid} than MEON. Equipped with a second-order pooling, DB-CNN trained on LIVE Challenge~\cite{ghadiyaram2016massive} is reasonably good at cross-distortion-scenario generalization. Not surprisingly, PQR, trained on KonIQ-10K with greater content and distortion diversity, generalizes much better to the rest databases than the one trained on BID.
%  PQR do not present any advantage over DB-CNN on databases including realistic distortions when they are trained on the large-scale KADID-10K and relatively small-scale CSIQ~\cite{larson2010most}, respectively, which demonstrates the challenge of generalizing the knowledge learned in the synthetic distortions to realistic distortions.
The most recent DNN-based method - PaQ-2-PiQ aims for local quality prediction, but only delivers reasonable performance on databases captured in the wild. \textcolor{black}{Relying on more sophisticated InceptionResNetV2 \cite{szegedy2016rethinking}, KonCept512 \cite{hosu2020koniq} presents promising results on three databases with realistic distortions.}  Enabled by the proposed training strategy, UNIQUE learns from multiple databases simultaneously, and \textcolor{black}{performs favorably against} all competing models on all six databases. It also shows competitive performance when comparing to state-of-the-art full-reference IQA models on synthetic databases.
% When UNIQUE is trained only on the combined training sets of KADID-10K and KonIQ-10K, it still performs well on the remaining four databases.
\begin{table*}[t]
  \centering
  \caption{Median results of UNIQUE trained with and without the hinge loss across ten sessions. The results in the last column are computed by the weighted average across all databases with the weight proportion to the number of images in each database}\label{tab:joint_evaluation}
  \begin{tabular}{l|ccccccc}
      \toprule
     {SRCC} & LIVE & CSIQ & KADID-10K & BID & LIVE Challenge & KonIQ-10K & Weighted\\
    \hline
            UNIQUE trained without the hinge loss  & 0.969 & 0.887 & 0.879 & 0.872 & 0.856 & 0.898 & 0.889 \\
        UNIQUE trained with the hinge loss & 0.969 & 0.902 & 0.878 & 0.858 & 0.854 & 0.896 & 0.888\\
    \midrule
    \color{black}{{PLCC}} & LIVE & CSIQ & KADID-10K & BID & LIVE Challenge & KonIQ-10K & Weighted\\
    \hline
        UNIQUE trained without the hinge loss  & 0.971 & 0.893 & 0.870 & 0.882 & 0.897 & 0.901 & 0.889 \\
        UNIQUE trained with the hinge loss & 0.968 & 0.927 & 0.876 & 0.873 & 0.890 & 0.901 & 0.892\\

    \midrule
     {Fidelity Loss} & LIVE & CSIQ & KADID-10K& BID & LIVE Challenge & KonIQ-10K & Weighted\\
    \hline
            UNIQUE trained without the hinge loss  & 0.131 & 0.086 & 0.048 & 0.058 & 0.015 & 0.026 & 0.041 \\
        UNIQUE trained with the hinge loss & 0.073 & 0.042 & 0.020 & 0.027 & 0.014 & 0.010 & 0.018\\
     \bottomrule
   \end{tabular}
\end{table*}

\begin{table}[t]
  \centering
  \caption{SRCC results on the four IQA databases under the cross-database setup. The subscripts ``s'' and ``r'' stand for models trained on KADID-10K~\cite{lin2019kadid} and KonIQ-10K~\cite{hosu2020koniq}, respectively. UNIQUE is trained on KADID-10K and KonIQ-10K simultaneously}\label{tab:cross}

  \begin{tabular}{l|cccc}
      \toprule
    Database & {LIVE} & {CSIQ} & {BID} & {LIVE Challenge}\\
    \hline
     NIQE & 0.906 & 0.627 & 0.459 & 0.449 \\
     ILNIQE & 0.898 & 0.815 & 0.496 & 0.439 \\
     dipIQ & \textbf{0.938} & 0.527 & 0.019 & 0.177 \\
     Ma19 & \textbf{0.919} & \textbf{0.915} & 0.295 & 0.330 \\
    \hline
     PQR$_s$ & 0.902 & 0.765 & 0.304 & 0.408 \\
     PQR$_r$ & 0.729 & 0.707 & 0.751 & \textbf{0.763} \\
     DB-CNN$_s$ & 0.916 & 0.751 & 0.602 & 0.531 \\
     DB-CNN$_r$ & 0.820 & 0.724 & \textbf{0.785} & 0.723 \\
    \hline
     UNIQUE & 0.917 & \textbf{0.830} & \textbf{0.783} & \textbf{0.786} \\
    %\midrule
    % PLCC & {LIVE} & {CSIQ} & {BID} & {LIVE Challenge}\\
    %\hline
    % Ma~\etal & 0.917 & 0.926 & 0.325 & 0.400 \\
    %\hline
    % PQR$_s$ & 0.908 & 0.826 & 0.326 & 0.516 \\
     \bottomrule
   \end{tabular}
\end{table}

\subsubsection{gMAD Competition Results}\label{subsubsec:gMAD}
gMAD competition~\cite{ma2020group} is a complementary methodology for IQA model comparison on large-scale databases without human annotations. Focusing on \textit{falsifying} perceptual models in the most efficient way, gMAD seeks pairs of images of similar quality predicted by one model, while being substantially different according to another model. To build the playground for gMAD, we combined all synthetically distorted images in the Waterloo Exploration Database~\cite{ma2017waterloo} with a corpus of realistically distorted images from SPAQ~\cite{fang2020perceptual}. \textcolor{black}{To quantify the competition results, we also conducted a subjective experiment following ITU-R BT.500~\cite{ITU}, and collected the MOS of each gMAD image. We gathered human data from $16$ subjects using  single stimulus continuous quality rating with the range of $[0, 100]$.} We first let UNIQUE compete against PQR~\cite{zeng2018blind} trained on the entire KonIQ-10K~\cite{hosu2020koniq} in Fig.~\ref{fig:gmad1}. The pair of images in (a) exhibit similar poor quality, which is in close agreement with UNIQUE. However, PQR favors the top JPEG compressed image, exposing its weaknesses at capturing synthetic distortions. When the roles of the two models are switched, UNIQUE consistently spots the failures of PQR (see (c) and (d)), suggesting that UNIQUE is better able to assess image quality in the laboratory and wild.

\begin{table*}[t]
  \centering
  \caption{Median SRCC results across ten sessions under different training strategies. MSE stands for the mean squared error}\label{tab:strategy_compare}
  \begin{tabular}{l|ccccccc}
      \toprule
     {SRCC} & LIVE& CSIQ & KADID-10K & BID & LIVE Challenge & KonIQ-10K & Weighted\\
     \hline
        Baseline (LIVE) & 0.951 & 0.721 & 0.475 & 0.632 & 0.502 & 0.688 & 0.596 \\
        Baseline (CSIQ) & 0.921 & 0.863 & 0.483 & 0.510 & 0.457 & 0.638 & 0.577 \\
        Baseline (KADID-10K) & 0.877 & 0.749 & 0.780 & 0.498 & 0.515 & 0.607 & 0.693 \\
        Baseline (BID) & 0.589 & 0.558 & 0.298 & 0.843 & 0.731 & 0.722 & 0.533 \\
        Baseline (LIVE Challenge) & 0.535 & 0.504 & 0.312 & 0.849 & 0.842 & 0.773 & 0.563 \\
        Baseline (KonIQ-10K) & 0.832 & 0.640 & 0.540 & 0.765 & 0.726 & 0.887 & 0.716 \\
    \hline
        Linear re-scaling + MSE (All databases) & 0.935 & 0.821 & {\bf 0.870} & 0.809 & 0.799 & 0.868 & 0.865\\
        Binary labeling + cross entropy (All databases) & {\bf 0.963} & {\bf 0.863} & 0.864 & {\bf 0.854} & {\bf 0.860} & {\bf 0.898} & {\bf 0.881}\\
        \color{black}{Continuous labeling + cross entropy (All databases)}  & 0.930 & 0.852 & 0.713 & 0.677 & 0.735 & 0.788 & 0.757 \\
    \hline
        %UIQE$_f$ (All databases)  & {\bf 0.969} & {\bf 0.887} & {\bf 0.879} & {\bf 0.872} & {\bf 0.856} & {\bf 0.898} & {\bf 0.889} \\
        UNIQUE (All databases)  & {\bf 0.969} & {\bf 0.902} & {\bf 0.878} & {\bf 0.858} & {\bf 0.854} & {\bf 0.896} & {\bf 0.888}\\
%    \midrule
%     {Fidelity Loss} $\downarrow$ & LIVE~\cite{sheikh2006statistical} & CSIQ~\cite{larson2010most} & KADID-10K~\cite{lin2019kadid} & BID~\cite{ciancio2011no} & LIVE Challenge~\cite{ghadiyaram2016massive} & KonIQ-10K~\cite{hosu2020koniq} & Weighted\\
%     \hline
%        UIQE$_f$ (All databases)  & 0.131 & 0.086 & 0.048 & 0.058 & 0.015 & 0.026 & 0.041 \\
%        UIQE (All databases)  & {\bf 0.073} & {\bf 0.042} & {\bf 0.020} & {\bf 0.027} & {\bf 0.014} & {\bf 0.010} & {\bf 0.018}\\
     \bottomrule
   \end{tabular}
\end{table*}

\begin{table*}[t]
 \footnotesize
  \centering
  \caption{Results of one-sided $t$-Test between our training method and  binary labeling. A value of $1$ indicates that our method is statistically superior to binary labeling; a value of $-1$ indicates the opposite; a value of $0$ indicates that the two methods are statistically indistinguishable}\label{tab:statistical_significance}
  \begin{tabular}{l|cccccc}
      \toprule
       & LIVE& CSIQ & KADID-10K & BID & LIVE Challenge & KonIQ-10K \\
    %  \hline
    %     Binary labeling & UNIQUE & -1 & -1 & -1 & -1 & 0 & 0 \\
    \hline
        Ours versus binary labelling & 1 & 1 & 1 & 1 & 0 & 0 \\
     \bottomrule
  \end{tabular}
\end{table*}

%complete
\begin{table*}[t]
  \centering
  \caption{Improved SRCC results of DB-CNN across ten sessions enabled by the proposed training strategy}\label{tab:dbcnn_uiqe}
  \begin{tabular}{l|cccccc}
      \toprule
     {SRCC} & LIVE~\cite{sheikh2006statistical} & CSIQ~\cite{larson2010most} & KADID-10K~\cite{lin2019kadid} & BID~\cite{ciancio2011no} & LIVE Challenge~\cite{ghadiyaram2016massive} & KonIQ-10K~\cite{hosu2020koniq}\\
     \hline
        %SRCC & 0.956 & 0.912 & 0.891 & 0.827 & 0.836 & 0.860 & 0.876\\
        DB-CNN (All databases) & 0.956 & 0.912 & 0.891 & 0.827 & 0.836 & 0.860\\
    \hline
    Improvements over DB-CNN (CSIQ) & 11.81\% & -- & 77.84\% & 151.37\% & 85.37\% & 72.34\%\\
    % \hline
    %  {PLCC} & LIVE~\cite{sheikh2006statistical} & CSIQ~\cite{larson2010most} & KADID-10K~\cite{lin2019kadid} & BID~\cite{ciancio2011no} & LIVE Challenge~\cite{ghadiyaram2016massive} & KonIQ-10K~\cite{hosu2020koniq}\\
    %     PLCC & 0.958 & 0.923 & 0.894 & 0.845 & 0.869 & 0.867 & 0.883\\
    \hline
    Improvements over DB-CNN (LIVE Challenge) & 32.22\% & 31.98\% & 82.58\% & 2.22\% & -- & 11.69\%\\
    %   Fidelity Loss $\downarrow$ & 0.073 & 0.035 & 0.019 & 0.030 & 0.019 & 0.017 & 0.021\\
     \bottomrule
   \end{tabular}
\end{table*}

We then let UNIQUE compete with DB-CNN~\cite{zhang2020blind}, which has demonstrated competitive gMAD performance against other DNN-based BIQA models~\cite{Ma2018End, bosse2016deep} on the Waterloo Exploration Database. In Fig.~\ref{fig:gmad2} (a) and (b), we observe that UNIQUE successfully survives from the attacks by DB-CNN, with images of similar quality according to human perception. DB-CNN~\cite{zhang2020blind} fails to penalize the top image in (a), which is severely degraded by a combination of out-of-focus blur and over-exposure. When UNIQUE serves as the attacker, it is able to falsify DB-CNN by finding the counterexamples in (c) and (d). This further validates that UNIQUE well aligns images across distortion scenarios in a learned perceptual scale.

\subsubsection{Uncertainty Estimation Results}\label{subsubsec:uncertainty_estimation}
UNIQUE can not only compute image quality scores, but also enable uncertainty quantification of such estimates. We tested the learned uncertainty function $\sigma_w(x)$ both quantitatively and qualitatively. To do so, we constructed a baseline version of UNIQUE, which was supervised by the fidelity loss only. That is, no direct supervision of $\sigma_w(x)$ is provided, and training
may suffer from the scaling ambiguity. In addition to SRCC and PLCC, we also adopted the fidelity loss as a more suitable quantitative measure as it takes into account the ground truth uncertainty when evaluation. Table~\ref{tab:joint_evaluation} shows the median results across ten sessions. Adding the hinge loss as a regularizer, UNIQUE presents a slightly inferior performance in terms of the weighted SRCC ($\approx0.1\%$), but is significantly better in terms of the fidelity loss ($>56\%$). This suggests that although the hinge loss may pose some constraints on quality estimation, it is helpful in regularizing the uncertainty learning of UNIQUE. We also draw the scatter plots of the learned uncertainty $\sigma_{w}(x)$ as a function of $f_{w}(x)$ in Fig.~\ref{fig:all_curves1}. With the hinge regularizer, the learned uncertainty on all databases exhibits human-like behavior, in that UNIQUE tends to assess images in the two ends of the quality range with higher confidence (\ie, lower uncertainty). In contrast, without the hinge regularizer, the learned uncertainty is less interpretable.
% In addition, the margin hinge loss serves as a constraint to bound both $f_{w}$ and $\sigma_{w}$ in reasonable ranges. Without such constraint, the range of $\sigma_{w}$ learned by UNIQUE$_{f}$ is erratic, \eg, the range of $\sigma_{w}$ of images in LIVE Challenge (see Fig.~\ref{fig:all_curves} (o)) is unreasonably smaller than that in other databases.

\subsubsection{Qualitative Results}\label{subsubsec:qualitative}
We conducted a qualitative analysis of UNIQUE by sampling images across different databases, as shown in  Fig.~\ref{fig:qualitative}. Although the proposed training strategy does not generate pairs of images from two different databases, the optimized UNIQUE is capable of aligning images from different databases in a perceptually meaningful way. In particular, synthetically distorted images with severe levels that may not encounter
in the real world receive the lowest quality scores, conforming to our  observations that the ranges of $f_w(x)$ in LIVE~\cite{sheikh2006statistical}, CSIQ~\cite{larson2010most} and KADID-10K~\cite{lin2019kadid} are relatively broader.

\subsection{Ablation Studies}
 \subsubsection{Performance in a Cross-Database Setting}\label{subsubsec:cross-database}
We tested UNIQUE in a more challenging cross-database setting. Specifically, we constructed another training set $\mathcal{D}'$ using image pairs sampled from the full KADID-10K~\cite{lin2019kadid} and KonIQ-10K~\cite{hosu2020koniq} databases, and re-trained it with the procedure described in \ref{subsec:exp_setup}. As comparison, we also re-trained two top-performing DNN-based methods - PQR~\cite{zeng2018blind} and DB-CNN~\cite{zhang2020blind} on the full KADID-10K~\cite{lin2019kadid} and KonIQ-10K~\cite{hosu2020koniq} databases. The full LIVE~\cite{sheikh2006statistical}, CSIQ~\cite{larson2010most}, BID~\cite{ciancio2011no}, and LIVE Challenge~\cite{ghadiyaram2016massive} were employed as the test sets. It is clear from Table~\ref{tab:cross} that UNIQUE achieves significantly better performance than the four knowledge-driven models and the two DNN-based models. The training image pairs from the two databases effectively provide mutual regularization, guiding UNIQUE
to a better local optimum. This experiment provides strong evidence that UNIQUE enabled by the proposed training strategy generalizes to both synthetic and realistic distortions.

\subsubsection{Performance of Different Training Strategies}\label{subsubsec:trainstrat}
The key idea of UNIQUE to meet the cross-distortion-scenario challenge is to train it on multiple databases. Here we compared different training strategies against the proposed one. We first treated BIQA as a standard regression problem, and trained six variants of UNIQUE on six databases separately using the mean squared error (MSE) as the loss. Next, we turned to exploit the idea of training BIQA models on multiple databases simultaneously. As discussed in Section~\ref{sec:intro}, we linearly re-scaled the MOSs to the same range of $[0, 100]$,  and re-trained the network using MSE. We also computed binary labels using MOSs: for an image pair $(x, y)$, the ground truth is $r=1$ if $u(x)\ge u(y)$ and $r=0$ otherwise. We re-trained the model using the cross entropy loss.  \textcolor{black}{ We further trained a variant of UNIQUE by replacing the fidelity loss with the cross-entropy loss. The results are listed in Table \ref{tab:strategy_compare}.} We find that regression-based models perform favorably on the training databases, present reasonable generalization to databases of similar distortion scenarios, but show incompetence in cross-distortion-scenario generalization. Training on multiple databases lead to a significant performance boost. \textcolor{black}{It is quite surprising that continuous labeling with the cross entropy loss offers the least improvement, even underperforming
linear re-scaling. This may be due to the instability of the cross entropy function when dealing with continuous annotations \cite{tsai2007frank}.} By contrast, binary labeling achieves better performance, but is still inferior to the proposed training strategy in terms of the weighted SRCC. \textcolor{black}{
We also conducted a $t$-test~\cite{Sheskin2004Handbook} on the SRCC values of all databases obtained from the ten sessions to examine whether the performance difference between our method and binary labeling is statistically significant. The null hypothesis is that the mean SRCC of our method is statistically indistinguishable from the mean SRCC of binary labeling with a confidence of $95\%$. From Table~\ref{tab:statistical_significance}, we observe that our training method delivers statistically superior performance on LIVE, CSIQ, KADID-10K, and BID, while is statistically on par with binary labeling on the remaining two databases.}
%

%Finally, we use the proposed method described in Sec.~\ref{sec:uiqe} to train the model on all databases. Compared with the model trained with binary labels, UIQE achieves consistently better performance on synthetically distorted databases and comparable results on realistically distorted databases, resulting in the best weighted average SRCC result.

%complete

%to do

%\begin{figure*}[t]
%    \centering
%    \captionsetup{justification=centering}
%    \subfloat[]{\includegraphics[width=0.33\textwidth]{figs/qualitative1}}
%    \subfloat[]{\includegraphics[width=0.33\textwidth]{figs/qualitative2}}
%    \subfloat[]{\includegraphics[width=0.33\textwidth]{figs/qualitative3}}
%  \caption{Qualitative examples. pmos and pstd mean predicted MOS and std, respectively. (a): Images with high objective perceptual quality. (b):  Images with middle objective perceptual quality. (c): Images with low objective perceptual quality. In each column, images are arranged from top to bottom according to the corresponding predicted quality in descending order. Images are cropped for better visibility.}
%\label{fig:qualitative}
%\end{figure*}

%to do

%to do

%to do

\subsubsection{Improving Existing BIQA Models}\label{subsubsec:improving}
The proposed training strategy is model-agnostic, meaning that it can be used to fine-tune existing differentiable BIQA models for improved performance. Here we implement this idea by applying the proposed training strategy to DB-CNN~\cite{zhang2020blind}. The SRCC results are shown in Table~\ref{tab:dbcnn_uiqe}.  We find that DB-CNN fine-tuned by the proposed training strategy achieves significantly better performance than the original versions trained on CSIQ and LIVE challenge, where the improvement can be as high as $151.37\%$. From this experiment, we may conclude that the improvement by the proposed training strategy is orthogonal to the network architecture design.

\section{Conclusion}
\label{sec:conclusion}
We have introduced a unified uncertainty-aware BIQA model - UNIQUE, and a method of training it on multiple IQA databases simultaneously. We also proposed an uncertainty regularizer, which enables direct supervision from the ground truth human uncertainty. UNIQUE is the first of its kind with superior cross-distortion-scenario generalization. We believe this performance improvements arise because of 1) the continuous ranking annotation that provides a more accurate supervisory signal, 2) the fidelity loss that assigns appropriate penalties to image pairs with different probabilities, and 3) the hinge regularizer that offers better statistical modeling of the uncertainty. In the future, we hope that the proposed learning strategy will become a standard solution for existing and next-generation BIQA models to meet the cross-distortion-scenario challenge. We also would like to explore the limits of the proposed learning strategy, towards \textit{universal} visual quality assessment of digital images and videos in various multimedia applications.

\section*{Acknowledgements}
The authors would like to thank Jingxian Huang for coordinating the subjective experiment and all subjects for participation.

\bibliographystyle{IEEEtran}
\bibliography{Weixia}

\end{document}